# Explainable Machine Learning for Categorical and Mixed Data with Lossless Visualization

Boris Kovalerchuk, Elijah McCoy
Dept. of Computer Science
Central Washington University,  USA
Boris.Kovalerchuk@cwu.edu, Elijah.McCoy@cwu.edu

*Abstract.* Building accurate and interpretable Machine Learning (ML) models for heterogeneous/mixed data is a long-standing challenge for  algorithms designed for numeric data. This work focuses on developing numeric coding schemes for non-numeric attributes for ML algorithms to support accurate and explainable ML models, methods for lossless visualization of n-D non-numeric categorical data  with visual rule discovery in these visualizations, and accurate and explainable ML models for categorical data.  This study proposes a classification of mixed data types and analyzes their important role in Machine Learning. It presents a toolkit for enforcing interpretability of all internal operations of ML algorithms on mixed data with a visual data exploration on mixed data. A new Sequential Rule Generation (SRG) algorithm for explainable rule generation with categorical data is proposed and successfully evaluated in multiple computational experiments. This work is one of the steps to the full scope ML algorithms for mixed data supported by lossless visualization of n-D data in General Line Coordinates beyond Parallel Coordinates.

1. INTRODUCTION

Many Machine Learning (ML) datasets contain **mixed/heterogeneous non-numeric data**, but multiple ML algorithms cannot discover models on such diverse data [31]. Mixed data include text, graphics, numeric and non-numeric data. Non-numeric data frequently take the form of ordinal (ordered) data, such as large, medium, and tiny, or nominal data with values like red, green, and blue.

The success of explainable ML algorithms for mixed data heavily depends on abilities of lossless visualization of multidimensional data and ML models. The latter enables end-users, who can contribute valuable domain knowledge that is missing from the training data, to create explainable ML models using **visual knowledge discovery** [13, 20].

Additionally, the interpretability and appeal of **visual machine learning models** are frequently higher for the end user than those of analytical ML models. This creates a new window of opportunity for developing and advancing the field of explainable ML.

It requires to visualize n-D data *without loss of n-D information*, because it is not known in advance which of this n-D information will be critical for discovering ML model in a visual form.

However, creating methods for **lossless visualization** for multiple data types is a long-standing challenge [1,10, 11], because it is more difficult than producing lossy visualizations of high dimensional data, which are less demanding on data preservation.

For instance, [33] focuses on existing lossy methods like 2-D projections and limited Star coordinates with visualizing only end points of the graphs making this visualization lossy relative to n-D data. In general, research on the lossless visualization of mixed nominal, ordinal, and numerical attributes and its contribution to the explainable machine learning is in the nascent stage.

Numeric coding is a popular method for adapting existing ML algorithms to non-numeric data [4-6]. If the features of these non-numeric data, such as value similarities, are corrupted by the coding, it can result in **non-explainable** ML models. For distance-based algorithms like kNN and others it happened with a common integer coding. Unfortunately, there is little focus on interpretability in the coding literature, even though it is a significant barrier to expanding the use of ML with mixed data in the domains with high error costs, like medicine.

Building Explainable Machine Learning (IML) models is a significant challenge for all types of data [21-27]. It requires understanding the *nature of the information* contained in the input data relative to the domain theory, and the nature of the knowledge that learning algorithms discover [18, 21]. It is also a *considerable computational challenge* especially for rule lists/sets [24], which is very evident for categorical data for large datasets with multiple value of each attribute. A *brute force* approach for categorical data is a computationally intractable problem with the exponential number of computations. A feasible approach to deal with this challenge is a **sequential hierarchical approach** that is presented in this paper based on the theory of monotone Boolean functions.

Often IML models are *unintelligible* being large and complex leading to the need to build smaller models [18, 27]. An extensive review of current approaches to build smaller explainable models as sets of rules is presented in [27]. Our approach differs from [27] in treating combinatorics of all possible rules. We use the computational advantages derived from the theory of monotone Boolean functions [28,30] with controlled levels of model precision and coverage to build an IML model as a set of explainable rules. We also do not require that a very complex explainable model is already available to be simplified as it is assumed in [27].

Most of the typical ML algorithms, like neural networks, SVM, kNN, Linear Discriminant Functions (LDF), CNN, and others are designed for quantitative numeric attributes such as pixel intensities in images. However, many machine learning data have a qualitative nature measured either in nominal or order scale, like good, bad, very good. Those attributes typically are encoded using one-hot key method which produces binary attributes [4].

The problem with one hot key is that it dramatically increases the sizes of the space. For instance, if an attribute has ten nominal values, one hot key will produce 10 binary attributes. Moreover, the interpretability of ML models such as kNN is questionable with one hot key. It counts the number of binary values equal or not equal mixed with differences between values of numeric attributes. It creates a *scaling problem* to make the total distances meaningful.

Moreover, many ML algorithms require distance metrics, which makes questionable the interpretability of those models on qualitative data. In contrast, we create the logical rules for qualitative data, which resolve such interpretability issue. This approach generates fully explainable and visualizable rules presented in this paper. The algorithm to generate such rules is denoted as **Sequential Rule Generation (SRG)** algorithm. This chapter presents several versions of this algorithm starting from the initial SRG0 algorithm.

Respectively, the main contribution of this chapter is in

(1) Integrating numeric **coding schemes** for non-numeric attributes for ML algorithms to support accurate and explainable ML models.
(2) Developing **accurate** and **explainable** ML models for categorical and mixed data.
(3) Developing methods for lossless **visualization** and **visual knowledge discovery** of n-D **non-numeric** categorical data and **visual rule discovery** in these visualizations.

The main innovation of this chapter in (1) is in the *tight match and control of knowledge* of input data types and operations of ML algorithms that can be allowed. The main innovation in (2) is in the *algorithms that cut out computations* using the theory of Monotone Boolean Functions. The main innovation in (3) is in *frequency-based visualization* and *visual knowledge discover*y of qualitative data in parallel coordinates. In contrast with [4] it is *lossless* without merging different values with equal frequencies including frequencies referencing the target attribute. This lossless visualization is a step toward the solution of the long-standing challenge of **lossless visualization** for multiple data types outlined above.

This paper is organized as follows. Section 2 proposes a classification of **mixed data types** and analyzes their important role in Machine Learning. Section 3 describes the proposed approach and the methodology for enforcing interpretability of internal operations of ML algorithms on mixed data. Section 4 presents different version of the **Sequential Rule Generation (SRG) algorithm** with a summary of verification experiments. Section 5 describes the proposed **visual exploration** approach and mixed data visualization methods. Section 6 summarizes the results and outlines the future work.

Appendixes A1-A10 detail experiments conducted with SRG algorithms on real-world data and Appendix A11 describes the toolkit developed.

## 2. HETEROGENEOUS DATA TYPES AND THEIR ROLE IN MACHINE LEARNING

In this section, mixed data types are categorized along with analyses of how important they are for machine learning. The discussion of various coding schemas for measurement data types is included.

### 2.1. Classification of mixed data types for machine learning

Heterogeneity/mixture of data is not limited by numeric, nominal, and ordinal data types that are **measurement data types** [16]. **Numeric** data are not homogeneous, but heterogeneous too. They have several measurement types such as **absolute, ratio**, **interval**, and **cyclical**. The strongest one (absolute) has the largest set of meaningful arithmetic relations and operations, which ML algorithms can conduct with them to produce explainable models.

This data type includes attributes that represent the counting the number of elements, which has unique not arbitrary 0 and unit 1. The next type is **ratio** data type, which has a unique zero, but not a unique unit, like Kelvin temperature with unique physical zero, but the unit is defined by convention. Other data types include **interval** data type (e.g., Celsius and Fahrenheit temperature where both zero and the unit are defined by convention) and values can be negative, **cyclical** (e.g., azimuth) [15] and others.

These data types can provide useful *arithmetic operations, distances, frequency distributions, distinct averages, standard deviation*, and other quantities that ML algorithms can use to create explainable models. However, the meaningful difference between two values in the cyclical azimuth attribute between $1°$ and $359°$ is $2°$ not 358 as it is for the attribute like weight. The **ordinal** data, in contrast, *forbid arithmetic operations* and only allow testing relations like $a \leqslant b$. Therefore, ML algorithms that perform these operations on ordinal data will result in models that cannot be explained.

Finally, only testing $a \neq b$ is meaningful for ML algorithms with **nominal** data. Other actions will result in unexplainable ML models. Graphs and texts are other examples of data heterogeneity, which are not given by a set of well-defined attributes. The attributes need to be made from them. We will call such data as **non-attribute-based data types**.

**Real-world physical modality data types**. Both height and weight are ratio attributes with physical zeros of weight and height, but arbitrary units and very different real-world physical meaning, modalities. Their sum and other mathematical operations, such as 3kg + 5m, are **meaningless**. This makes popular **linear** machine learning models non-explainable for data of different modalities and with mutual dependencies [18, 22].

In contrast, ML models with data of the same modality type as temperature in time series can be explainable.

Thus, each dataset is characterized by the several categories of **data types**: 1) type of **measurement**, (2) **attribute/non-attribute**, and (3) type of **modality.** A dataset is **heterogeneous** or **mixed** if it contains data of various such types. Differences in data types have a significant impact on the precision and interpretability of ML models. The ML algorithm likely will produce incorrect and unintelligible models when used with data types for which it was not intended.

While the effects of measurement and attribute/non-attribute data types [4-6] are widely acknowledged, the effects of **physical modality** are less recognized in the ML literature. This has led to several unsupported claims like that linear models are always comprehensible/interpretable, which was critically analyzed in [18].

The process of **embedding**, which focuses on converting non-numeric data like graphs and texts [17] into quantitative data (strings of numbers, vectors), is the subject of numerous recent studies. Although this technique is widely used and aims to maintain the structure of the original data, it has a serious flaw. The interpretability of such vectors is unclear or nonexistent, while these models are frequently quite accurate.

**Encoding** individual attributes based on their measurement data types [4-6] allows building explainable models simpler than with embedding from mixed data. To reduce the attribute space and to make attributes more relevant to the ML task, the data preparation stage frequently encodes the original attributes to other attributes by many-to-one mappings. It results in **loss** of some information and can corrupt the data type.

Example. The absolute numeric data in [0,100] interval can be converted to ordinal data, where code 1 is for interval [0,50), code 2 for [50,70) and code 3 for [70,100]. The resulting codes will be $cd(a)=1$, $cd(b)$, and $cd(c)$, respectively, for the values $a=10$, $b=60$, and $c=100$. In the original absolute data, $c=10a$ and $b=6a$, but in codes $cd(b)=2cd(a)$ and $cd(c)=3cd(a)$, if we treat codes as numeric data of the same measurement type as original data.

This example demonstrates that the data type cannot be automatically expanded to a modified attribute. The validity of the code ratios $cd(b)=2cd(a)$ and $cd(c)=3cd(a)$ in the domain should be confirmed by a domain expert. Otherwise, the ML model that uses these codes is not explainable in that domain.

We cannot consider codes as a numeric data type without such confirmation from a domain expert; instead, we must treat them as ordinal data, where computing ratios is both forbidden and meaningless. The disparities $cd(b)-cd(a)=2-1=1$ and $cd(c)-cd(a)=3-1=2$ also have little significance in the absence of equivalent confirmation. Therefore, distance-based ML algorithms like kNN will result in unexplainable models in this situation. Unfortunately, the focus of the kNN literature reviewed in [31] is out of the interpretability issues.

Calculating the dissimilarity between each pair of nominal attribute values in relation to the target attribute for the ML classification problem is one method for enriching nominal attributes. This can be achieved by constructing the adaptive dissimilarity matrices and minimizing an error function on the training samples while accounting for attribute correlation [7]. This method called ADM generalizes the Value Difference Metric (VDM) [8]. A similar idea is presented in [5] for SVM. Such approach improved the ML model accuracy and interpretations of the relationships among the different values of nominal attributes [7] for some datasets from UCI ML repository. However, for the mushroom data [9], which we consider, the explainable C4.5 Decision Tree (DT) algorithm produced the lowest error of 0.4% reported in [7] without using ADM or VDM. Because DTs can overlook some superior possibilities when examining each attribute only one at a time during node splitting, this lowest error is still not zero [7].

The main challenge with ADM is that the error function is computed using a machine learning algorithm on training data. Another ML algorithm will lead to a different result. Next, if the ML algorithm is unexplainable, then the ADM outcomes will likewise be unexplainable too.

The radial basis function (RDF) classifier was used in [7] as the ML algorithm. It sums up the exp functions of the distances. The interpretability of it for the mushroom task is not clear, in contrast with decision trees or logical rules. The result with RDF needs to be independently confirmed by the domain knowledge, which was done in [7] for the odor mushroom attribute. The coding that is explainable in the domain and established without using the target attribute does not require such domain confirmation afterward.

*2.2. Coding of measurement data types*

The major coding techniques for nominal, ordinal, and other attributes used in machine learning [4-6] are briefly discussed below, along with their relations to interpretability. The development of these techniques to accommodate knowledge discovery in multidimensional data utilizing visualization will be presented in section 4.

**One Hot Encoding** maps a nominal attribute to a binary vector, where 1 denotes the existence of the provided attribute value and 0 denotes the absence of that value of the attribute. It greatly lengthens the run time and space dimension for the ML algorithms. This coding is meaningful for nominal attributes because it preserves the equal Hamming distances between any two values of the nominal attribute.

The **label encoding** simply assigns numbers from 1 to $k$ to $k$ values of the nominal attribute. It does not preserve the equal distance for values of the nominal attribute. Thus, it is not explainable, which is especially applicable to the ML algorithm that exploit the differences of distances for classification like kNN. Such coding can be used for other techniques, such as decision trees (DTs), which do not use n-D distances, but it may result in less effective DTs than with alternative encodings including another numbering of values of the attribute.

**Hashing** maps categorical attributes to vectors in n-D space, where the distance between two vectors is roughly preserved. Compared to One Hot Encoding, the final dimension is significantly lower [4]. This is a subcategory of a more comprehensive embedding notion for various kinds of heterogeneous data. Not every ML algorithm can benefit from hashing. Preserving the structure for nominal attributes means all equal distances between values. If the ML algorithm exploits the differences of the distances, then it cannot benefit from such hashing, where all distances are the same. Establishing meaningful distances for ordinal data require additional information, which may not be available. It limits the use of the hashing method for these data. A hashing space can be **not explainable**, which is critical.

**Ordinal encoding** is applicable to the **ordinal data** like very short, short, medium, tall, very tall. It differs from the label coding by the fact that now numbering cannot be arbitrary, but it must follow the order of the values, e.g., 1 for very short, 2 for short, 3, for medium, 4 for tall, and 5 for very tall. The explainable operations with ordinal data are limited by **less or equal relation** and should not include arithmetic operations. This limits application of many existing ML algorithms that use arithmetic operations.

Below we summarize statistics-based coding methods [4]. These methods are applicable to many data types but can assign the same code to different values of the attribute making this coding **lossy**. We resolve this issue for visualization in Section 4.

    **Frequency Encoding** assigns codes to values of the attribute by their frequency. It allows exploring the link between attribute and the values of the target variable too.

    **Mean Encoding** or Target Encoding computes the mean of the number of times the value of the attribute appears in the target class in the two-class classification task

    **Probability Ratio Encoding** for each value uses ratio $P(1)/P(0)$ of the frequency of this value of the attribute for class 1 to the frequency for class 0 as codes.

    **James-Stein estimator** assigns as a code a weighted average of the mean target value for the observed feature value and the mean target value (regardless of the feature value).

3. APPROACH AND METHODOLOGY

This section describes the proposed approach and the methodology for enforcing *interpretability of all internal operations of ML algorithms on mixed data*. The Toolkit that implements this approach is outlined in the appendix A11.

**Enforcing interpretability of all internal operations of ML algorithms**. The main idea behind the proposed methodology is *to produce explainable models, where the internal operations of the ML algorithm must be comprehensible/explainable*. It is a significant restriction for models that deal with mixed data. For instance, the algorithm cannot perform any action for nominal data other than verifying whether two values are equal or not because all other operations cannot be justified for nominal data.

This leads to algorithms, which discover **logical rules** like (1)-(3) or similar:

$$\text{If } [(x_1\text{=a} \ \& \ x_2\text{=b}) \lor (x_3\text{=c})] \ \& \ x_4 \neq d \ \& (\neg \ x_5 \neq e) \text{ Then } \mathbf{x} \in \text{class } C_1 \qquad (1)$$

$$\text{If } [(x_1\text{=a} \ \& \ x_2\text{=b}) \lor (x_3\text{=c})] \ \& \ x_4 \neq d \ \& \ (\neg x_5 \neq e) \text{ Then } \mathbf{x} \in \text{class } C_1 \text{else } \mathbf{x} \in \text{class } C_2 \qquad (2)$$

$$\text{If } [(x_1\text{=}y_1 \ \& \ x_2\text{=}y_2) \lor (x_3\text{=c})] \ \& \ x_4 \neq y_4 \ \& \ (\neg x_5 \neq e) \ \& \ \mathbf{x} \in \text{class } C_1 \text{ Then } \mathbf{y} \in \text{class } C_1 \qquad (3)$$

for nominal n-D points $\mathbf{x}=(x_1,x_2,\ldots x_n)$, $\mathbf{y}=(y_1,y_2,\ldots y_n)$. Thus, the rules can include only logical operations $\&, \lor, \neg$ and tests if values are equal or not for one or more n-D points.

Even decision trees which are traditionally considered as explainable ML algorithms are **not formally explainable** for the nominal data because they check $\leqslant$ relation, which is prohibited for nominal data. Coding of nominal values by integers (known as label encoding) allows technically to apply decision trees to nominal data. However, randomness of this coding leads to different decision trees and does not guarantee finding the most accurate model. Next, the produced DT must be converted to logical rules with elimination of all $\leqslant$ relations and their thresholds to make it fully explainable.

The advantage of ML algorithms, which build **logical rules** is that they are **data type universal**, i.e., can produce models that include *heterogenous* data of **all types** that are expressed in the propositional or the First order Logic (FoL) [14]. Therefore, this work focuses on logical rules. Above rules (1) and (2) are **propositional** and rule (3) is an example of **FoL rules**, because it includes the FoL clause $x_4 \neq y_4$.

**Addressing all heterogeneous data types**. The actual number of data types in ML tasks with mixed data is much greater than what is usually listed. Consider a nominal data type for which we cannot say that **a** is closer to **b** than to **c**, like occupations with categories: doctors, engineers, and teachers. Next, we can add a nurse to this list and can continue to call it the nominal data type. Alternatively, we can say that doctors and nurses are closer to each other than to engineers and teachers. Then, occupations are *not nominal anymore* because more relations have meaning. Similarly, we can add technicians and teaching assistants. This creates more relations, where engineers are closer to technicians and teachers are closer to teaching assistants than to other occupations. To treat these occupations as a nominal data type we can create groups: (1) doctors and nurses, (2) engineers and technicians, and (3) teachers and teaching assistants and assign codes to these groups.

While grouping allows us to go back to the nominal data, we lose important similarity information about occupations. The resulting ML model on groups can be less accurate than without grouping. Instead of grouping we can assign the following codes: nurse (1), doctor (2), technician (5), engineer (6), teaching assistant (10), teacher (11) with limiting the set of operations and relations that are considered as explainable and allowable. For instance, relation,

$$c(\text{doctor})-c(\text{nurse}) < c(\text{teacher})-c(\text{engineer}) \qquad (4)$$

is allowed. Here, c(.) is an integer code for the occupation, e.g., c(nurse)=1. Thus, the ML algorithm can use relation (4), but not (5):

$$c(doctor)-c(engineer) < c(teacher)-c(engineer) \qquad (5)$$

This example shows that to get an explainable ML model with non-numeric data like occupations we can use a numeric coding of them, but we need to **limit operations** and **relations**, which can be conducted with these numbers by ML algorithms to ensure that algorithms will produce explainable models. In summary, this example shows that the design of ML algorithms for mixed data needs to address a wide variety of data types. The **relational ML algorithms** [14] fit well this task for mixed data.

**Building explainable models based on the explainable atoms (hyperblocks).** The next important concept in our methodology for explainable ML on mixed data is the concept of the n-D data hyperblocks (HBs). Some HBs serve as *explainable data atoms*.

A **numeric hyperblock** (hyperrectangle, n-orthotope) is a set of numeric n-D points $\{\mathbf{x}=(x_1,x_2,\ldots,x_n)\}$ with **center** in n-D point $\mathbf{c}=(c_1,c_2,\ldots,c_n)$ and *lengths* $\mathbf{L}=(L_1, L_2,\ldots, L_n)$ such that

$$\forall i \in I_u \, \| x_i - c_i \| \leqslant L_i/2 \qquad (6)$$

This definition assumes that attributes are **numeric** and the difference between values is **meaningful**. This means that data are of the interval data type at least. For heterogenous data we need another definition which will include different data types. We first define the hyper-block for ordinal data and then for nominal data.

An **ordinal hyperblock** (hyperrectangle, n-orthotope) is a set of ordinal n-D points $\{\mathbf{x}=(x_1,x_2,\ldots,x_n)\}$ with *edge ordinal* n-D points $\mathbf{s}=(s_1,s_2,\ldots,s_n)$ $\mathbf{e}=(e_1,e_2,\ldots,e_n)$ such that

$$\forall i \in I_o \, s_i \leqslant x_i \leqslant e_i \qquad (7)$$

For instance, if $X_i$ has values,1,2,3,4,5 and $s_i$=2, and $e_i$=4 then n-D points with values 2,3, or 4 of $X_i$ will be in this hyper-block

A **nominal hyperblock** is a set of nominal n-D points $\{\mathbf{x}=(x_1,x_2,\ldots,x_n)\}$ such that

$$\forall i \in I_n \, x_i \in Q_i \qquad (8)$$

where $Q_i$ is a subset of values of attribute $X_i$. For instance, $X_i$ ={doctor, teacher, engineer} and $Q_i$ ={doctor, teacher}.

A **heterogeneous hyperblock** is a set of n-D points $\{\mathbf{x}=(x_1,x_2,\ldots,x_n)\}$ such that
$\forall i \in I_u \, \| x_i - c_i \| \leqslant L_i/2$ for all numeric attributes
$\forall i \in I_o \, s_i \leqslant x_i \leqslant e_i$ for all ordinal attributes and
$\forall i \in I_n \, x_i \in Q_i$ for all nominal attributes.

All numeric HBs are **explainable** for data with numeric attributes of different modality types like temperature and blood pressure. HBs are explainable because: (i) the distance in (6) is defined within each numeric attribute, where it is meaningful and (ii) do not include arithmetic operations between heterogenous attributes, which are not meaningful, but only combine them with the meaningful logic operation. Thus, each such HB is a logical model, e.g., a HB contains all cases with temperature in [35,37] interval and blood pressure within [100,120] interval. Similarly, the hyperblocks for ordinal and nominal data are explainable. As a result, this hyperblock for heterogeneous data is also explainable.

Next larger explainable heterogeneous datasets can be formed from smaller HBs. These small HBs serve as **explainable atoms** for such larger datasets. We can formulate a stronger **conjecture** that **all** explainable heterogeneous datasets with numeric, ordinal, and nominal attributes can be formed from HBs.

A hyperblocks is called a **pure HB** if it contains only cases of a single class. For every n-D point **x** it is possible to find a single class HB if there is no n-D point **y=x** that belongs for another class. Another advantage of numeric HBs is that they can be **visualized losslessly**, which is shown in [3].

The practice of coding the values of a nominal attribute by consecutive integers 1,2,…,n known as label encoding [4] is *contradictory* from stating that (1) it can **always** be used to encode nominal attributes without any limitation to stating that (2) this coding should **never** be used because it is not explainable. In fact, it should not be used in ML algorithms, which conduct arithmetic operations with these codes like subtraction, squaring and so on. It is common in ML algorithms that use distances between n-D points like kNN. The use of this coding with the ML algorithms that do not make arithmetic operations with values of nominal attributes does not create interpretability problems. The logical algorithms, which we outlined above belong to this category. Another advantage of creating and visualizing logical rules is that often they are used as an efficient **tool to explain deep learning** and other black box models by mimicking behavior of these models.

4. SEQUENTIAL RULE GENERATION (SRG) ALGORITHMS

In this section we present the **Sequential Rule Generation** (**SRG**) algorithm for finding rules in categorical data analytically. After such rules will be discovered they can be visualized in parallel coordinates similarly to shown in Section 4. This approach generates fully explainable and visualizable rules.

*4.1 Rule generation process*

We consider **qualitative data**, which are represented by **nominal** and **order** attributes without obvious numeric values. Below first we focus on nominal attributes. Discovering rules with these data is challenging computationally. A brute force algorithm to discover rules on the available data will need to test all rules like:

$$R_1: \text{If } x_1=3 \ \& \ x_4=1_i \ \& \ x_{10}=4 \text{ then } \mathbf{x} \text{ belongs to class } C \tag{1}$$

for all possible subsets of *n* attributes, all possible values of these attributes and on all n-D points (cases/samples) of available data. The number of these subsets is $2^n$ and is growing exponentially with *n*. For each subset of attributes, we need to test several rules. Assume that in rule (1) attribute $x_1$ has three values (1,2,3), $x_4$ has two values (1,2) and $x_{10}$ has 4 values (1,2,3,4), then $k_1k_2k_3=3\cdot2\cdot4=24$ such rules must be tested. If each attribute has only 2 values, then each subset with *q* attributes will have $2^q$ rules. Next each rule must be tested on all *m* given n-D points. This shows the exponential complexity of the brute force algorithm. We cannot generate and test all rules in this way for large *n, m*, and *q*. Therefore, Sequential Rule Generation algorithm generates rules sequentially and filters out not promising rules without full testing them.

The filtering methods in the Sequential Rule Generation algorithm is based on the **principle of monotone Boolean functions** [28,30] as follows. Consider rule $R_1$ from (1) and rule $R_2$, which contains rule $R_1$ as a subrule, where $R_2$ is generated by adding an extra clause $x_6=3$:

$R_2$: If $x_1=3$ & $x_4= 1_i$ & $x_{10}=4$ & & $x_6=3$ then **x** belongs to class *C*  (2)

If $R_1$ has low coverage of cases, then $R_2$ will also have low coverage because an additional requirement $x_6=3$ will at most keep the number of cases that satisfy $R_2$ the same as for $R_1$ but most typically it will be less than $R_1$. If we already *filtered out* $R_1$ as covering too few cases, then we can filter out $R_2$ without testing it on the data. Similarly, if $R_1$ has greater precision than $R_2$, then $R_2$ will also have the same or greater precision. If we already satisfied with $R_1$ precision (e.g., 100%), then we can omit testing $R_2$ because it is more complex and keeping $R_1$ is sufficient for the selected level of precision. Thus, SRG algorithm allows cutting out brute force generation of rules by using **monotonicity** properties, which filter out not promising rules. It is done with the techniques of monotone Boolean functions [28,30].

To cut out the brute force computations further the SRG algorithms splits the attributes to **groups** and for each group the search for rules is conducted based on monotonicity approach described above. In our experiments, groups contained 3-5 attributes. For instance, in one of the experiments with the mushroom data, which contain 22 attributes, the groups were formed as follows: Group 1: $x_1, x_2, x_3$, Group 2: $x_4, x_5, x_6,…$,Group 7: $x_{19}, x_{20}, x_{21}, x_{22}$.

According to the design of SRG algorithm it runs at the several levels of **thresholds** that the user sets up. We experimented with 3 level of precision: 75%. 85%, 95% and with different levels of coverage. These thresholds limit only the low margin of the rule quality, but do not limit their upper level. Therefore, we also computed the actual precision and coverage reached for the poisonous class at each level.

The **monotone Boolean process** for *n*=3 is presented below. We build three Hansel chains [28], which cover all 8 triples that represent subsets of attributes to be used to generate rules. For instance, the triple (110) means that only attributes $x_1$ and $x_2$ will be used to build rules, but $x_3$ is excluded.

**Example**. Figs. 1 and 2 show the following chains:

Chain 1: (010), (110); Chain 2: (100), (101); Chain 3: (000),(001),(011),(111)

First, we check triple (010) on Chain 1 if any rule for this triple of attributes satisfies the requirement for 75% precision and 0.5% coverage. This triple means that only attribute $x_2$ is used, because only the second bit in (010) is 1. If such rule would be found then no other triples on chain 1 is tested. The illustrative example in Fig. 2 shows that no such rule is found for attribute $x_2$. Therefore, we need to test the next triple (110) on chain 1, i.e., rules with attributes $x_1$ and $x_2$.

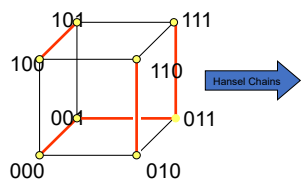 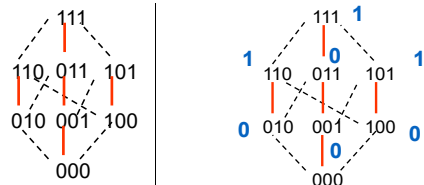

Fig. 1. 3-D Hansel chains unlabeled        Fig. 2. Labeled Hansel chains

Fig. 2 shows that such rule was found, and the triple (110) is labeled by 1. It immediately implies that (111) on chain 3 is excluded from testing because it is greater than (110). The process moves to test the first triple (100) on chain 2 and fails to find a rule that satisfies the requirements of 75% precision and 0.5% coverage in $x_1$ only, which corresponds to (100) triple. Then we successfully check the next triple (101) on chain 2. Then the process moves to chain 3 until all remaining triples on this chain are tested starting from (001). In Fig. 2. it will require to test (001) and (011).

Here we illustrated the process starting from the bottom of each chain, Alternative processes that start from the top of each chain or from the middle of each chain are also implemented and depending on the data set the cutting of computations can be more efficient with one of these options. Starting from the longest chain also can change the number of tests required. The domain knowledge can help selecting one of these processes. If the domain expert expects that 2 or more of the attributes are required for rules to be successful out of three attributes then starting from the top will be more beneficial to cut computations than starting from the bottom of chains.

*4.2 General Algorithm design and SRG0 Algorithm*

The SRG algorithm attempts to discover a set of rules for class $C$ such that each rule $R$ captures some properties of a **single class**:

if $R(\mathbf{x})=1$ then case **x** belongs to class $C$

with the following desirable properties: (1) a high number of cases that $R$ predicts/covers, $N_{covered}$, (2) a high number of cases that $R$ predicts correctly, $N_{correct}$, (3) a small set of rules, and (4) simple rules.

Below we will use the following concepts for every rule $R$, which predicts class $C$:

**Recall** is $N_{correct}/N$, , where $N_{correct}$ is the number of cases **x** that satisfy rule $R$ and *belong* to class $C$, $N$ is the *total number* of cases of class $C$.

**Precision** is $N_{correct}/(N_{correct}+N_{incorrect})$, where $N_{incorrect}$ is the number of cases **x** that satisfy rule $R$, but do *not* belong to class $C$.

**Coverage** is ( $N_{correct}+N_{incorrec}$

**Examples**. Let for rule $R_2$: If $x_1$=3 & $x_4$= $1_i$ & $x_{10}$=4 & & $x_6$=3 Then **x** $\in C$  the value $N=100$ (the number of cases from class $C$ which satisfy the if-part of this rule). If $N_{correct}$=80, $N_{incorrect}$=10 then we get recall=80/100, precision=80/(80+10), and coverage=(80+10)/100. Note that coverage can be greater than 1 when $N_{incorrect}$ is large, e.g., if 50 cases of another class would satisfy the rule in this example. If the rule predicts cases of two classes, then $N$ will be the number of cases from these two classes. The values of $N_{correct}$ and $N_{incorrect}$ also will be adjusted for two classes. The advantage of discovering rules that are limited to a subset of cases of a **single class** is that in many tasks we do not have enough knowledge to discover rules that will classify every possible case of all classes correctly and with high confidence. So, it is better to leave those cases unclassified, than to predict them incorrectly or with a low confidence.

While the SRG algorithm discovers only single class rules, datasets contain cases of several classes. In the **two-class classification**, the rule $R`$ for the opposite class $C`$ can be generated from $R$ by reversing it, $R`(\mathbf{x})=1 \Leftrightarrow R(\mathbf{x})=0$, i.e., If $R`(\mathbf{x})=1$ then **x** belongs to class $C`$. Respectively, the **coverage and precision** of rule $R`$ heavily depend on these properties of rule R. This shows the importance of maximizing the coverage and precision of all cases of class $C$ by rule $R$ to be able to predict accurately cases of the opposite class. In many situations, precision, recall, and coverage contradict each other and the **tradeoff** between them is needed. This chapter presents different versions of the Sequential Rule Generation algorithm with different tradeoffs and different options to form groups of attributes (sequentially, randomly, by an expert, and by guidance of existing ML model) to decrease the search time.

The complexity of a rule and set of rules are defined as the number of the **base clauses** in the rules relative to the number of cases covered. A base clause is a base predicate in the rule like $x_i$=5, or $x_i \neq 5$.

**Complexity of Rule** $R$ = (Number of base clauses in $R$)/ (Number of cases covered by rule $R$).
**Complexity of a set of rules** = (Number of clauses in the set)/ (Number of cases covered by the set of rules).

The number of cases covered by a set of rules can be computed as: (1) a sum of cases covered by each rule independently or (2) dependently counting each case only once, while it can be covered by multiples rules. The selection of this method depends on the goal of the complexity analysis. To measure the rule confidence (1) can be more informative, but to measure coverage (2) can be more informative.

Example. Rule R1 has a complexity of 1/1000 and rule R2 has a complexity of 100/1000 then it is intuitively clear that R1 is less complex than R2 because the last one requires 100 times more base clauses to cover the same number of cases as the first rule R1.

The **base algorithm** (**SRG0**) consists of the following steps:
   (1) Form groups of attributes sequentially, e.g., group 1: ($x_1,x_2,x_3$), group 2: ($x_4,x_5,x_6,x_7$) and so on. In the current experiments each groups contained 3-5 attributes per group typically.
   (2) Generate a set of rules in each group by using Monotone Boolean Functions (MBFs) with computing their precision and coverage
   (3) Sort rules by the number of cases they cover in decreasing order, e.g., $R_5$:1000 cases, $R_3$: 900 cases, $R_2$:800 cases, $R_7$:700;
   (4) Search for rules in the sorted list with 100% precision.
   (5) If such rules found (e.g., $R_3$ with 900 cases and $R_7$ with 800 cases) record

Thus, SRG0 algorithm is for the "ideal" data, where rules with 100% precision exist.

*4.3 Algorithm SRG1 with rule overlap analysis*

Algorithm SRG1 improves rule discovery by analyzing the overlap of the rules with each other on underlying data. If rules do not overlap, then they are fully complementary. If rules overlap, then we have more options to modify and **combine** them. In addition, rule overlap is useful for increasing the **confidence in prediction**. For example, consider case **x** covered by three rules. These rules use different attributes and different their values, but still predict that case **x** belongs to the same class *C*. This gives a *higher confidence* to the prediction that **x** belongs to class *C* like poisonous mushroom than if we would have only one rule. This is in line with the **voting** approach that is actively and successfully used in machine learning, especially in the random forest algorithm and other ensemble-based methods.

SRG1 algorithm specifies how rules are selected from generated rules for the selected class *C*, which we call a **target class**. This process **orders** generated rules by the number of cases in class *C* that are covered by each individual rule but are **not covered** by already selected rules. It uses the advantage of selecting rules with the highest number of class *C* cases that are not covered by already selected rules to cover class *C*.

Now consider a situation when one *rule dominates and covers majority* of the cases of the given class with high precision. This rule must be highlighted for the deep analysis by the domain expert/analyst. The reason is that it is unlikely that this rule is spurious and overfitting data, especially if it is short containing only few attributes and few values of those attributes. This rule indicates that attributes and their values are **most informative** ones. So, it allows to focus attention on these attributes to limit later data collection by those attributes and measuring only their values.

If other rules cover only a few cases, then they have a **low generalizability** and **low trust/confidence**. The modification of them by increasing their coverage will increase the confidence in rules and will avoid data overfitting and memorization. This is another way to look at the confidence of the rules. Thus, all rules that cover many cases increase the confidence in the set of rules and those rules should be preferred rules to be included instead of rules that cover only a few remaining cases not covered by the dominant rule.

The **SRG1 algorithm** consists of the following steps:
(1) Form groups of attributes sequentially, e.g., group 1: $(x_1,x_2,x_3)$, group 2: $(x_4,x_5,x_6,x_7)$ and so on.
(2) Generate a set of rules in each group by using Monotone Boolean Functions (MBF) with analysis of the *overlap* of rules in the process of rule generation.
(3) Sort rules by the number of cases they cover in decreasing order, e.g., $R_5$:1000 cases, $R_3$: 900 cases, $R_2$:800 cases, $R_7$:700.
(4) Determine the first candidate rule $R_{candidate}$ and record the value of $N_{notCovered}$ for it.
(5) Select all other rules, denoted as *OtherRules,* with the same value of $N_{notCovered}$.
(6) If any rule in *OtherRules* has greater $N_{covered}$ and greater precision $P$, select it and repeat the process for finding better rules.

Thus, SRG1 algorithm selects rules with less overlap or with more cases that are
not covered by prior rules. It differs from SRG0 in steps (4)-(6).

### 4.4. Algorithm SRG2 with complementary rules for precision improvement

As shown above, the mushroom dataset allowed to achieve full coverage of the target value "poisonous" with 100% precision by requiring 100% precision in the rule discovery. Above the SRG0 algorithm required all rules to have 100% precision. This however is **not feasible** for datasets that have no rules with 100% precision. Therefore, in this section we present a rule selection process that first allows selecting less precise rules and then update the rules to increase precision. It is done by discovering rules that predict an *opposite class,* which is eatable for the mushroom data. We will also call it **non-target class**. The algorithm records all incorrectly predicted cases by initially selected rules, and then generates new rules to predict the opposite class on these data. For example, if 900 cases are incorrectly predicted as class 1 that are in fact in class 2, then the algorithm would generate rules to predict **class 2** using those 900 cases.

A produced algorithm is denoted as SRG2 algorithm. It assumes a prior run of algorithms SRG0 and SRG1 on the same data with grouping attributes. After that the steps of the **SRG2 algorithm** are as follows:

(1) Choose **most frequent attributes** to be used to generate rules in Step (2). The most frequent attributes are derived from a prior run of algorithms SRG0 and SRG1 on the same data with grouping attributes. The attribute is considered as a most frequent if it appeared in 50% of the rules that each group generated.

(2) Sort chosen attributes by the percentage of appearance per attribute grouping and form groups based on this sorting with the first group with the most frequent attributes.

(3) Generate all rules for the **target class** (e.g., poisonous) with chosen attributes and select the rules that lead to the highest coverage and precision while recording incorrectly predicted cases.

(4) Generate all rules for the **non-target class** (e.g., eatable) using the incorrectly predicted cases. These rules are called **complementary** rules.

(5) Add rules that predict the non-target class cases to the final selected rules if the rules increase overall precision and meet the following requirements,
> Requirement 1: Rule must cover previous incorrectly predicted cases.
> Requirement 2: Rule must not incorrectly predict target class cases.

(6) Combining rules for two classes (see the next section for details).

### 4.5. Algorithm SRG3 based on 30 randomly generated triples of attributes

Version 3 of Sequential Rule Generation algorithm, SRG3, differs from versions SRG0-SRG2 in the selection attributes to the groups. In the previous versions groups are created **sequentially**, like $(x_1,x_2,x_3)$, $(x_4,x_5,x_6)$ and so on. In SRG3 groups of three attributes are formed **randomly** and 30 random such triples are created.

Accordingly, we have versions of SRG3 based on the SRG1 and SRG2 as follows. **SRG3 based on SRG1** uses 30 randomly created triples of attributes. **SRG3 based on SRG2** uses most frequent attributes from 30 randomly created triples in the run of SRG3 based on SRG1.

The exploration of algorithm SRG3 based on algorithm SRG1 allows to see if it could achieve 100% coverage and 100% precision of class $C_1$ with a small set of rules without relying on the original attribute order or any other specific attribute groupings. This is done in the experiment 4 on Mushroom data.

The exploration based on the SRG2 allows to see how the frequency of attributes in the rules impacts the quality of the rules. Those **most frequent attributes** are selected from the results of running SRG3 based on SRG1. This is done in the experiment 5 on the same Mushroom data.

### *4.6. Algorithm SRG4 based on expert selected groups*

The next version of the sequential rule generation algorithm, denoted as SRG4 differs from the prior versions in that the groups of the attributes are generated by the **domain expert**. For the mushroom data it was a biologist. This was done to see if the mushroom data attributes could be grouped using expert knowledge to produce high coverage and high precision rules.

The main difficulty when creating attribute groupings is that there is usually little knowledge of how the attributes relate to each other. By allowing the expert biologist to create attribute groupings, the hope is that this difficulty will decrease.

### *4.7. Algorithm SRG5 based on prior suceesfull attributes*

The idea of the next version of the sequential rule generation algorithm, denoted as **SRG5** is selecting attributes that have been successful in the other models on the same data. This is a form on the **knowledge transfer**.

The sources of these attributes can be attributes that are revealed in previous high precision rules on these data. The expectation is that it will allow the rule generation process to generate and select less rules while keeping a high precision and total coverage of the target class. To shorten computation time this to the different groups.

### *4.8. Summary of 10 experiments with SRG Algorithms*

The detailed results of 10 experiments with SRG algorithm on the mushroom data from [9] are presented in appendices A1-A10. Table 1 provides a summary of these results. The target class of rules in all experiments is poisonous.This table allows us to conclude that SRG algorithm is capable to discover useful and compact interpretable rules from the mushroom data with high precision and coverage reaching 100% in both. This indicates the feasibility to discover such rules on other data sets with qualitative attributes, which needs additional exploration.

Analysis of these results shows that selecting a specific SRG algorithm for the task at hand depends on a combination of several properties of the task and data. If the goal is to get 100% precision and coverage then the SRD0 can be applied. While it happened with the mushroom data, we cannot expect it for all possible data.

For such data SRG will be more appropriate with abilities to set up a desired level of precision and coverage, like 75%, 85%, or 95%. If additional knowledge is available like expert grouping of attributes, or attributes that have been successful in prior models on the same data then SRG45 are preferable.

Table 1. Summary of ten experiments with SRG algorithm on mushroom data.

| # | Algorithm | Description | Rules | Precision % | Coverage % | Highlight |
|---|---|---|---|---|---|---|
| 1 | SRG0 | Sequential attribute triples | 7 | 100 | 100 | 96.94% coverage of the best rule $R_1$ |
| 2 | SRG1 | rule overlap analysis | 7 | 100 | 100 | The non-overlap of rules $R_1$ and $R_6$, increases coverage |
| 3 | SRG2 | Complimentary rules, 95% precision threefold | 13 | [95.2,100] | 100 | Rule with largest coverage $R_1$, 98.47% coverage and 95.2% precision |
| 4 | SRG3 | 30 random attribute groups | 7 | 100 | 100 | 96.94% coverage of the best rule $R_1$ |
| 5 | SRG3 | 13 most frequent attributes | 7 | 100 | 100 | 96.94% coverage of the best rule R1. Did not reduce the number of rules below 7. |
| 6 | SRG3 | Sequential attribute triples,10-fold cross validation | 4 | 100 | 100 | All 10-fold cross validation folds acquire 100% accuracy and precision. |
| 7 | SRG3 | 30 random attribute triples (4 times), 10-fold cross validation | 4-5 | 100 | 100 | All 10-fold cross validation folds acquire 100% accuracy and precision. Complexity of rule sets [13/3525, 16/3520]. |
| 8 | SRG4 | Expert selected groups. 95% precision threshold | 3 | 100 | 99.81 | Smallest number of rules with a good coverage. |
| 9 | SRG5 | 7 successful attributes and groups. 95% precision threshold | 7 | 100 | 100 | Did not reduce the number of rules below 7. |
| 10 | SRG5 | Attributes and groups from [29,32]. 95% precision threshold | 4 | 100 | 100 | Lower complexity of rules. Complexity 0.002 vs. 0.0027 for [29,32] with 11 clauses vs 13 clauses for [29,32]. |

Table 2 summarizes the appropriate situations and goals for SRG versions.

Table 2. Situations and goals appropriate for versions of SRG algorithm.

| Algorithm | Situation | Goal |
|---|---|---|
| SRG0 | Given order of attributes is good for attribute grouping | 100% precision and coverage. |
| SRG1 | Given order of attributes is good for attribute grouping | Short set of rules |
| SRG2 | Given order of attributes is good for attribute grouping | Precision and coverage above thresholds |
| SRG3 | No information on good attribute grouping. | 100% precision and coverage or short set of rules |
| SRG 4 | Expert selected attribute groups | Precision and coverage above thresholds |
| SRG 5 | Prior successful attributes in rules are given | Precision and coverage above thresholds |

## 5. Mixed Multidimensional Data Visualization and visual rule discovery

In the previous sections we demonstrated abilities to discover rules analytically with an appropriate coding of categorical data. This section presents methods to visualize mixed multidimensional data and rules along with rule discovery in parallel coordinates.

The rules produced in Experiment 10 are presented in Appendix A10 in detail. These rules are less complex than CR rules from [29,32]. Our rules use 11 clauses covering all poisonous cases with duplication (5428 cases, complexity 11/5428=**0.002**). CR rules use 13 clauses to cover all poisonous cases with less duplication (4780 cases, complexity 13/4780=**0.0027**).

**R$_1$:** $[(x_5=3) \lor (x_5=4) \lor (x_5=5) \lor (x_5=6) \lor (x_5=8) \lor (x_5=9)] \Rightarrow \mathbf{x} \in C_1$
**R$_2$:** $[(x_{20}=5)] \Rightarrow \mathbf{x} \in C_1$
**R$_3$:** $[(x_{12}=3) \,\&\, (x_{21}=5)] \Rightarrow \mathbf{x} \in C_1$
**R$_4$:** $[(x_8 \neq 1) \,\&\, (x_{21}=2)] \Rightarrow \mathbf{x} \in C_1$

**CR Rules** [29,32]
**CR$_1$:** $[(x_5=3) \lor (x_5=4) \lor (x_5=5) \lor (x_5=6) \lor (x_5=8) \lor (x_5=9)] \Rightarrow \mathbf{x} \in C_1$,
**CR$_2$:** $[(x_{20}=5)] \Rightarrow \mathbf{x} \in C_1$
**CR$_{31}$:** $(x_8=2) \,\&\, (x_{12}=3) \Rightarrow \mathbf{x} \in C_1$
**CR$_{32}$:** $(x_8=2) \,\&\, (x_{12}=2) \Rightarrow \mathbf{x} \in C_1$
**CR$_{33}$:** $(x_8=2) \,\&\, (x_{21}=2) \Rightarrow \mathbf{x} \in C_1$

Our rules have been discovered using extensive analytical computations by machine learning SRG5 algorithm described in Section 4. Similarly, CR rules also have been discovered using extensive computations by other machine learning methods. These computations are hidden from the user to be able to trust the produced rules. A user should be able to get some kind of justification of that the rules correct. Visualization is an efficient method to help users to understand and trust rules.

These rules are visualized in Fig. 3 in parallel coordinates. They contain only 5 attributes out of 22 attributes providing 100 precision and coverage of all mushroom poisonous cases. The visualization of rules in Fig. 3 is quite simple and can be understood by the end users easily.

Rules R$_1$ and R$_2$ are the same as rules CR$_1$ and CR$_2$. In Fig 3, visually rule R$_3$ is simpler than a joint rule of rules CR$_{31}$ and CR$_{32}$ shown in blue. In contrast the rule R$_3$ visually looks a more complex than rule CR$_{33}$ in gray, while they have the same computational complexity, R4 checks $x_8 \neq 1$ and CR$_{33}$ checks $x_8=2$. This shows the deficiency of parallel coordinate visualization, which does not allow compactly visualize the negation property $x_8 \neq 1$. It requires showing all values which are not equal to 1 on $X_8$.

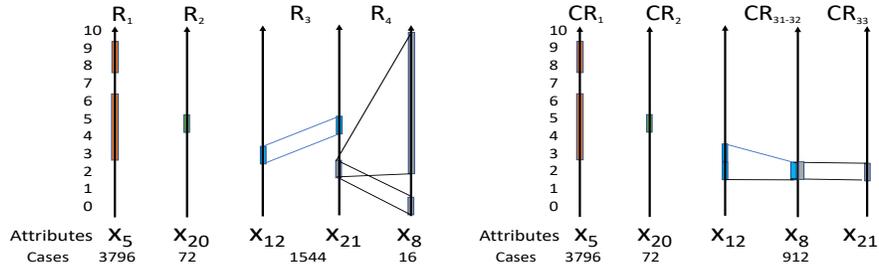

(a) Visualization of rules from Experiment 10.   (b) Visualization of rules from [29,32].
Fig. 3. Visualization of rules generated in Experiment 10 and from [29,32] for poisonous mushroom.

It is relatively easy to provide to the user a visual confirmation of these rules by plotting all poisonous cases in these rules in Fig. 3. For rule $R_1$ it will locate each case that satisfy $R_1$ as a point in the red area on attribute $X_5$. For $R_2$ it will be points in the in the green area on attribute $X_{20}$ with value 5. For $R_3$ it will be all lines connecting value 3 in $X_{12}$ with value 5 in $X_{21}$ and for rule $R_4$ it will be all lines connecting value 2 of $X_{21}$ with all other values in $X_8$ but 1.

However, drawing in Fig. 3 all mushroom cases fully showing their values not only in 1-2 attributes but 5 or all 22 attributes will produce significant overlap and occlusion of cases. It creates difficulties to analyze cases fully, to observe discovered patterns that rules present and to discover new patterns visually. Solving this task requires visual knowledge discovery (VKD) methods that we present in this section.

Fig. 4 shows all mushroom data visualized in parallel coordinates in VisCanvas [2,3], where nominal attributes are encoded by binary attributes using the Data Type Editor of our Toolkit described in Appendix 11 resulted in over 70 attributes. The uninformative attributes with the same values for all cases are omitted leaving 60 attributes shown in Fig. 4.

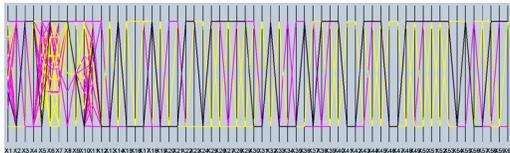

Fig. 4. Binary encoded mushroom data in parallel coordinates.

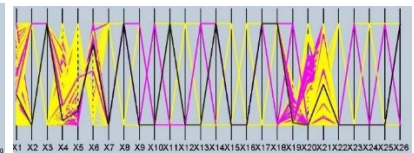

Fig. 5. Mixed census income data in parallel coordinates after meaningful coding.

These 60 attributes from 22 original attributes demonstrate deficiency of such coding for visualization. The data are quite hard to look at since there are so many attributes. Moreover, in general, binary attributes do not fit well to be visualized in parallel coordinates because binary attributes do not have much variability and do not fill the area between 0 and 1 in each coordinate. As a result, many lines cover each other making different classes practically undistinguishable as was shown before in [10].

Fig. 5 shows visualization of the Census income dataset from the UCI ML repository [9] in parallel coordinates. It was selected for its size of 48842 cases of mixture of 14 integer and categorical attributes. Conversion several nominal attributes to binary attributes produced total 26 attributes. Binary coding of nominal attributes requires domain knowledge to make it meaningful, which is easier when attributes are a part of the common knowledge like in the Census data.

Figs. 4-6 show the deficiency of visualization of data with multiple nominal attributes by converting to binary attributes. Fig. 6 visualizes the Teaching Assistant (TA) evaluation dataset from the UCI ML repository [9], which is a mix of categorical and integer data. In these data values of some attributes are grouped and converted to binary attributes using the Data Type Editor. The cases are colored according to the values of the attribute X9, which has over 20 values. It shows high overlap of cases making visual discovering of patterns difficult. Therefore, below we present alternative methods.

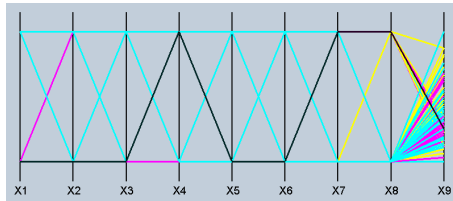 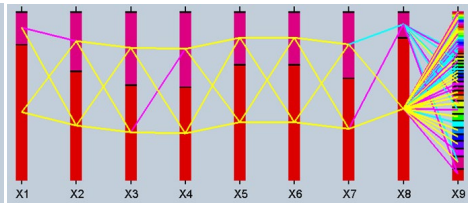

Fig. 6. TA data in Parallel coordinates.     Fig. 7. TA data in adapted parallel sets

**Frequency based visualization of nominal attributes**. Fig. 7 shows the visualization of the same TA dataset in parallel coordinates where bars are sized by the **frequency** of occurrence of the respective nominal values. The user interface in VisCanvas allows a user to generate this frequency-based visualization. This visualization is similar to the **parallel sets** visualization [12].

The important difference is that the lines going from one attribute to another do not have the same width as the bar on the attribute. Use **thin** lines making visualization less occluded. The frequencies of values of X9 also have been computed and visualized by the height of the bars.

The bars are ordered according to their frequencies in the descending order with bars of the greatest frequency at the bottom of X9. This visualization is more informative than shown in Fig. 6, which helps to solve the occlusion problem that we see in Fig. 6. Also, a user can select a specific color scheme for the nominal blocks.

**Reference frequency visualization.** The frequency visualizations shown in Fig. 7 adapted from parallel sets [12] have an important deficiency for machine learning tasks. They are *not related to the classes,* the heigh of the bar for a given value of the attribute is based on the *frequency* of this value *itself* in the dataset, which contain cases of several classes. Below we present visualizations that are more relevant to machine learning. This method computes the number of bars, and their heights based on the **relations** of these values with values of another **reference attribute** $X_t$ (e.g., target attribute).

The coloring of lines and bars is based on the values of the reference attribute/classes. For instance, reference attribute can have 00 cases with $x_i$=a can contain 10 cases with $x_t$=0, 70 cases with $x_t$=1, 12 cases with $x_t$=2, and 8 cases with $x_t$=3. Here $x_t$=1 is a dominant value of $X_i$.

The bars for all non-dominant values can be joined to a single bar (see grey bars in Fig. 8). It visually emphasizes the dominant value of the target attribute/class. In Fig. 8 the portion of each bar is colored by the dominant class (magenta or blue) and the non-dominant part is grey, the black horizontal lines separate bars.

In addition, wider lines allow a user to see the larger frequency of cases between bars. In attribute $X_9$ (class size/number of students) all smaller frequencies under a threshold are put in one block at the top. In comparison with Fig. 6, it is now easier to see how many lines are going to each bar and understand dominance of classes.

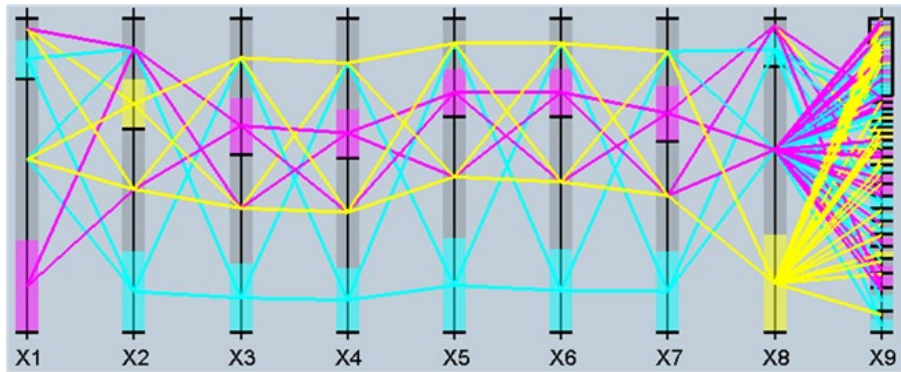
Fig. 8. Visualization with class colors and dominant frequency for TA dataset with weighted lines.

Figs. 9 and 10 illustrate advantages of reference frequency-based visualization on the mushroom classification dataset with class colors. Fig. 9 shows dominant class frequency with all bars, while Fig. 10 shows only bars of high purity (⩾80%).

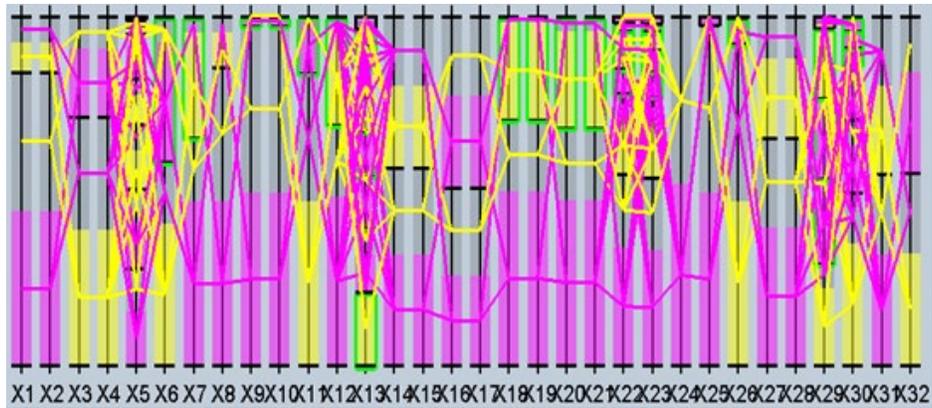
Fig. 9. Mushroom dataset with class colors and dominant frequency with all bars.

The frequency approach implemented in these visualizations differs from frequency methods listed in [4]. The **Frequency Encoding** in [4] converts textual data into numeric data by assigning the frequency of that value as its code. In this coding if two values get the same frequency, say, 0.3, then 0.3 will be used as code for both values making them indistinguishable. In our frequency visualization these two values will have their own bars of length 0.3 each. So, they will not collapse to a single bar and the information will be preserved. To distinguish our frequency encoding from described in [4] we will call our frequency encoding as a **visual frequency encoding**. The **Mean Encoding**/Target Encoding [4] has the same issue. It can produce the equal codes and glue values, while our visualization avoids the information loss.

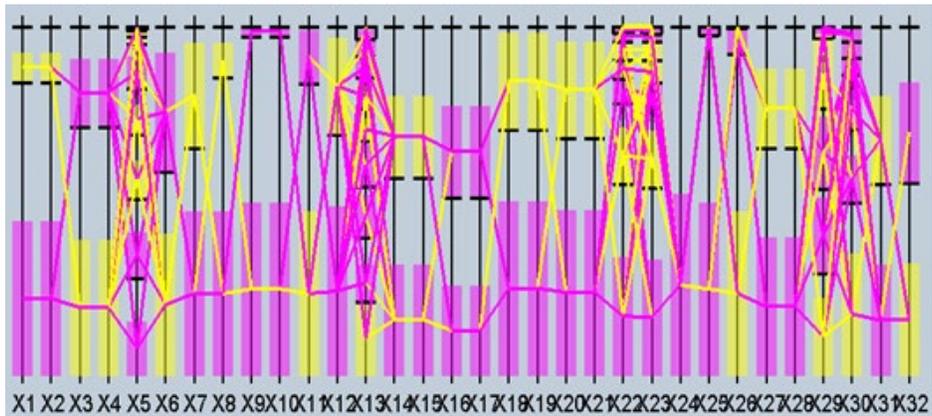
Fig. 10. Mushroom dataset with class colors with bars above 80% purity.

To help a user to get more information from the visualization their linguistic descriptions are automatically generated (Fig. 11). It describes blocks/bars with purity of 80% or higher and attributes with small frequency blocks/bars.

```
X1, block, 1 has a total frequency of 84
X2, block, 1 has a total frequency of 84
X5 has a small frequency block.
X6, block, 2 has a purity of 81
X7, block, 2 has a purity of 85
```

Fig.11. Dominant blocks/bars on Mushroom dataset.

Fig. 11 describes a few attributes of the mushroom dataset visualized in Fig. 10. On this dataset many attributes have small blocks with purity above 80% purity. To emphasize larger blocks the limit on block size of 10% was introduced in addition to 80% of purity.

In Fig. 8, attribute $X_9$ was not very useful in discovering rules. In contrast, Fig. 12 shows the TA dataset with splitting values of attribute $X_9$ (class size) into 4 groups. Now it shows the pattern that TAs rated the best (blue blocks and lines) were dominant in the first and the fourth groups. The worst rated TAs were dominant in the second group and the average TAs were dominant in the third group.

**Flipping attributes** allows making visual patterns clearer and VisCanvas supports it. Fig. 13 shows flipping (negating) some attributes in the TA dataset. All attributes are normalized to [0,1] and flipping creates 1-x for the attribute x. As a result, blue cases concentrate at the bottom, magenta cases are in the middle and yellow cases split between the top and the middle.

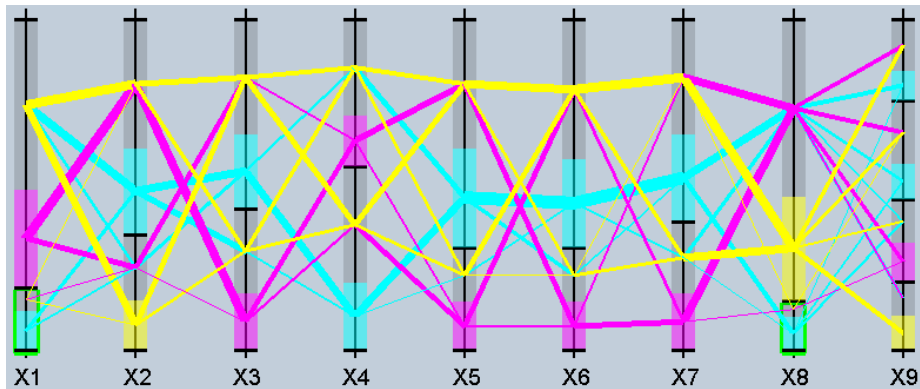

Fig. 12. TA dataset with blocks sorted by purity and wider frequent cases.

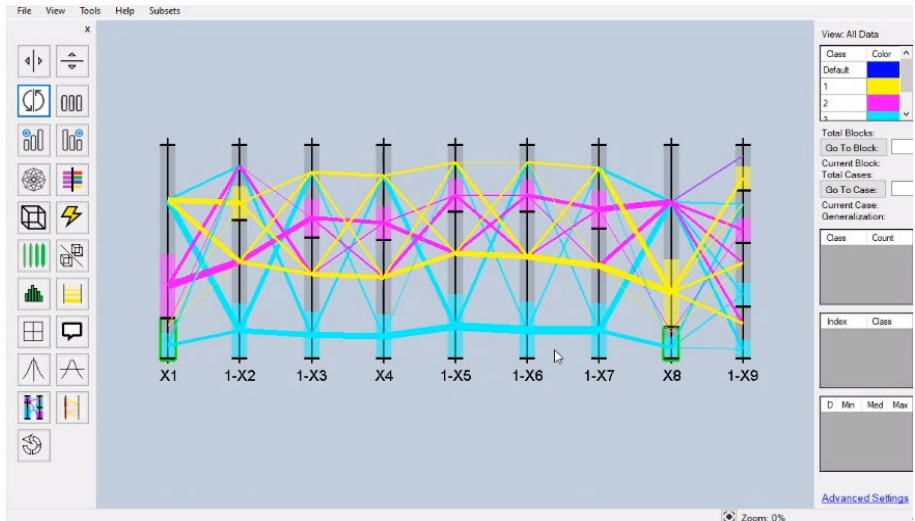

Fig. 13. TA dataset with flipped attributes to simplify visuals.

**Reordering attributes** is another option in VisCanvas to make patters clearer. Fig. 14 shows the mushroom data with reordered attributes, where the attributes with the puree blocks are on the left and the less pure blocks are on the right. It makes visual patters clearer. A user can change the purity threshold for more distilled visual patterns.

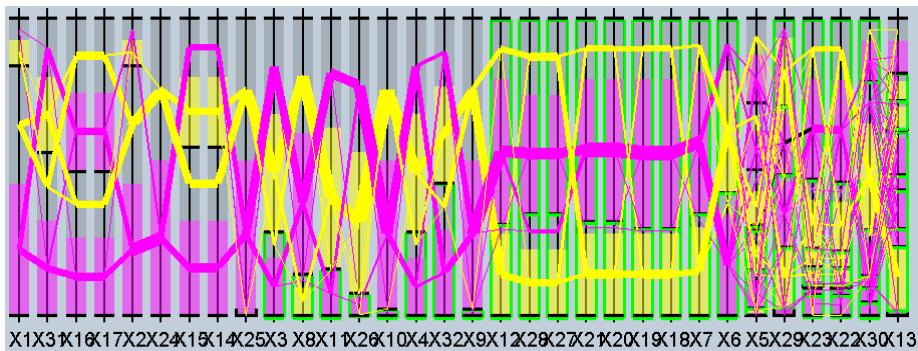

Fig. 14. Mushroom data with attributes ordered by purity of blocks decreasing from left to right and wider frequent cases.

**Grouping, relocating blocks**, **and sorting by color** is another way to make patterns simpler and clearer. Some datasets have attributes with a few large blocks and many small blocks. Often these small blocks are of high purity and are next to the large blocks, which making them hard to see as in Fig. 14 on the right. Therefore, the *smaller blocks* (under 20%) are moved to the top of those attributes (see Fig. 15). Fig. 15 also shows the results of *sorting* the mushroom data by *color* and putting the yellow blocks on the top. Here the attributes are sorted by the number of purity blocks.

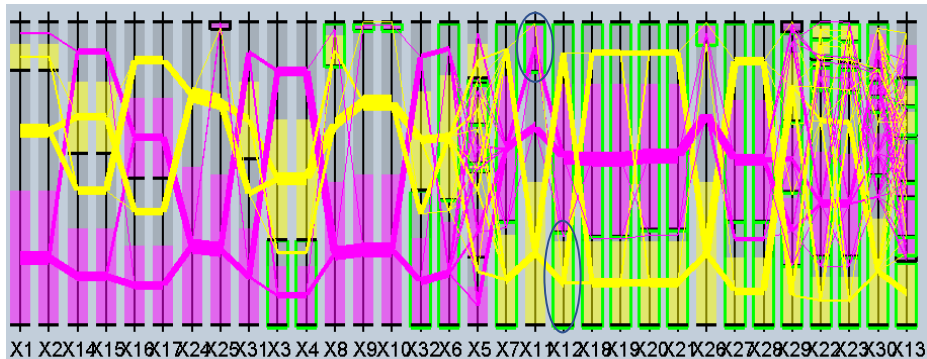

Fig. 15. Mushroom dataset sorted by color.

**Visual rule generation**. Such strong visual patterns in the data allow to make classification rules like

If case **a** is in magenta block in X11 & is NOT
in yellow block X12 then **a** is in yellow class        (7)

These blocks are outlined by the black ovals in Fig. 15. These magenta and yellow boxes have green frames indicating their high purity. This rule is an example of explainable rules that end users can find themselves by such visual knowledge discovery process. The comparison of Fig. 15 and Fig. 4 with the same data shows the benefit of the toolkit. Extracting rules like (7) is impossible in Fig. 4.

**The benefits of visual knowledge discovery** derived from visualization in Fig. 13 are described below for the TA dataset. This dataset is a collection of attributes of students who were rated 1 to 3 on their effectiveness of being a teaching assistant. The value 3 is a 'great' rating, 2 is an 'average' rating, and 1 is a 'bad' rating. In Fig. 13, 3 is blue, 2 is yellow, and 1 is magenta. After flipping X2-X7, and X9 we can see a clear pattern in the bottom of Fig. 13 that leads to a teaching assistant being great.

The attribute X1 shows a purity of over 60% dominant class 3 (great) in the bottom block and a dominant class of class 1(bad) in the top block. This attribute represents a Boolean value of whether the teaching assistant was a native English speaker or not where the top block is false, and the bottom block is true. This means that a teaching assistant is more likely to be rated as 'great' if they are native English speakers.

The first block in X2 is dominantly blue with most "great" TAs going there. It represents the value 0 meaning that a majority of student's label "great" had instructors from either group 1, 2, or 3. The upper block in the ovals with magenta dominant block is much less pure. Similarly, in X3 most of the "great" teaching assistants either had instructor in group 1 or group 4. Next, value 1 of the blue block in X4 means that the "great" teaching assistants came from group 1 or 3.

The analysis of attributes X5, X6, and X7 shows that "worst" rated TAs were assisting in classes from group 4 since the first block value is one. The attribute X8 represents whether the semester was a normal semester or a summer semester. The bottom block of this coordinate is dominantly blue. Its value 2 means that most "great" teaching assistants were helping in classes during summer semesters. Fig. 13 shows that the bottom two blocks are dominantly blue in attribute X9 (class size). This means that "great" teaching assistants dominantly helped in classes with 3 to 19 students or 37 to 66 students.

6. CONCLUSION AND FUTURE WORK

The focus of this study has been on developing: (1) numeric **coding schemes** for non-numeric attributes for ML algorithms to support accurate and explainable ML models, (2) **accurate** and **explainable** ML models for categorical data, and (3) methods for lossless **visualization** of n-D **non-numeric** categorical data with **visual rule discovery** in these visualizations. For this we proposed a classification of **mixed data types** and analyzes their important role in Machine Learning, an approach, and a toolkit for enforcing interpretability of all internal operations of ML algorithms on heterogeneous data with a **visual exploration** approach on mixed data. These methods have been demonstrated on Mushroom, Census income, and Teaching Assistants evaluation datasets. For (2) we proposed different versions of the **Sequential Rule Generation (SRG) algorithm** for explainable ML with categorical data. These versions offer options to group attributes (sequential, random, expert based, and by knowledge transfer from successful models).

These version of SRG algorithms have been successfully evaluated in multiple experiments on the benchmark Mushroom dataset. In experiments the different versions of SRG algorithm have been able to produce rules that accurately classify all mushroom data. The comparison with the rules generated in [29] had shown advantages of SRG generated rules relative to rules produced in [29]. SRG rules reached the same accuracy with less complexity. The versions of the SRG algorithm varied in the ways how the groups of attributes are formed, types of rules generated and how the selection of priority rules. The major issue, which we discovered in this study is computational limitations to explore the groups with the larger number of attributes. Further experiments are needed to test SRG algorithms, which will allow to discover other limitation of current versions of SRG and to develop new versions of SRG algorithm which will fit different datasets.

This study provides a user with the adaptation of mixed data types and their coding schemas for lossless visualization of multidimensional mixed data in parallel coordinates. The developed experimental Toolkit combines the Data Types Editor and VisCanvas system for mixed multidimensional data visualization and explainable rule discovery. It is available on GitHub [2]. It supports (a) numeric *coding schemes* for non-numeric attributes for *explainable ML* with mixed data, (b) *lossless visualization* of n-D *non-numeric* data, and (c) *visual rule discovery* in these visualizations, and (d) analytical rule discovery with Sequential Rule Generation algorithms. The future work is developing new *full scope ML algorithms* for mixed data integrated that will generalize SRG algorithms with lossless visualization of n-D heterogeneous data and other types of General Line Coordinates beyond Parallel Coordinates.

APPENDIX

*A1. Experiment 1 with algorithm SRG0 for discovering rules on Mushroom with sequential triples*

In this experiment we tested SRG0 algorithm for discovering rules on Mushroom data [9] to solve a two-class classification problem (poisonous - eatable). We found rules with **100% precision** for the *target class $C_1$ -- poisonous mushrooms.* One of these rules is a rule reported in the literature [29]. Below we present discovered rules R1-R7 using notation of the case $\mathbf{x}=(x_1,x_2,…,x_n)$. If any of these rules is true, then $\mathbf{x}$ belongs to class $C_1$.

**Rules R1-R7**
**R1:** $[(x_5=3) \vee (x_5=4) \vee (x_5=5) \vee (x_5=6) \vee (x_5=8) \vee (x_5=9)] \Rightarrow \mathbf{x} \in C_1$
**R2:** $[(x_9=6) \vee (x_9=3)] \Rightarrow \mathbf{x} \in C_1$
**R3:** $[(x_{19}=2) \& (x_{20}=8) \& (x_{21} \neq 2) \& (x_{22} \neq 2)] \Rightarrow \mathbf{x} \in C_1$
**R4**: $[(x_{15}=3) \vee (x_{15}=2) \vee (x_{15}=9)] \Rightarrow \mathbf{x} \in C_1$
**R5**: $[(x_{19} \neq 2) \& (x_{20} \neq 6) \& (x_{21}=5) \& (x_{22}=1)] \Rightarrow \mathbf{x} \in C_1$
**R6**: $[(x_{19}=6) \& (x_{20}=5) \& (x_{21} \neq 1) \& (x_{22}=!6)] \Rightarrow \mathbf{x} \in C_1$
**R7**: $[(x_{20}=8) \& (x_{21}=2) \& (x_{22}=!6)] \Rightarrow \mathbf{x} \in C_1$

Table A1 presents characteristics of discovered 7 rules, R1-R7, which **cover 100%** of the mushroom data for the "poisonous" class with **100% precision**. The total number of cases in the poisonous $N_{total}$ class is 3916 cases. The rules were with attributes broken up into six sequential triples and one group with 4 attributes: $x_{19}, x_{20}, x_{21}, x_{22}$. These 6 groups are $(x_1, x_2, x_3)$, $(x_4, x_5, x_6)$, $(x_7, x_8, x_9)$, $(x_{10}, x_{11}, x_{12})$, $(x_{13}, x_{14}, x_{35})$, $(x_{16}, x_{17}, x_{18})$.

Table A1. Characteristics of discovered rules R1-R7.

| *Characteristic* | R1 | R2 | R3 | R4 | R5 | R6 | R7 |
|---|---|---|---|---|---|---|---|
| Precision, % | 100 | 100 | 100 | 100 | 100 | 100 | 100 |
| Coverage, % | 96.94 | 44.74 | 49.34 | 12.56 | 9.91 | 1.84 | 1.33 |
| Total cases predicted, $N_{covered}$ | 3796 | 1752 | 1184 | 492 | 388 | 72 | 52 |
| Correctly predicted cases, $N_{correct}$ | 3796 | 1752 | 1184 | 492 | 388 | 72 | 52 |
| Misclassified cases | 0 | 0 | 0 | 0 | 0 | 0 | 0 |

*A2. Experiment 2 with SRG1 algorithm for discovering rules on Mushroom data with rule overlap minimization*

The analysis of rules presented in the previous section shows that, Rule $R_1$ covers 96.94% of the cases with target value "poisonous", which means that the other 6 rules add to the coverage only 3.04% of these cases (118 cases). For this reason, to get a better understanding of these rules we calculated the overlap between cases classified by different rules. See Tables A2 and A3. In these tables, the overlap, OL, is the total number of cases that are in the intersection of rules $R_i$ and $R_j$, $OL(R_i,R_j)=|Cases(R_i) \cap Cases(R_j)|$, the overlap percentage is $OL(R_i,R_j)$ divided by the total number of cases in the union of cases covered by both rules, $OL(R_i,R_j) / |cases(R_i) \cup cases(R_j)|$. Table A2 shows the relations of rule $R_1$ with other rules and Table A3 of remaining rules with each other.

Table A2. Overlap between dominant rule $R_1$ and other rules.

| | | $R_2$ | $R_3$ | $R_4$ | $R_5$ | $R_6$ | $R_7$ |
|---|---|---|---|---|---|---|---|
| | Cases covered | 1752 | 1184 | 492 | 388 | 72 | 52 |
| $R_1$ 3796 cases | Overlap, % | 45.24% | 30.09% | 12.25% | 9.19% | 0% | 0.94% |
| | Total cases | 3820 | 3828 | 3820 | 3832 | 3868 | 3812 |
| | Overlap cases | 1728 | 1152 | 468 | 352 | 0 | 36 |
| | Added cases | 24 | 32 | 24 | 36 | 72 | 16 |
| | Added cases, % | 20.00% | 26.67% | 20.00% | 30.00% | 60.00% | 13.33% |

For instance, the analysis of the relations between rules $R_1$ and $R_2$ show that together they cover 3820 cases of poisonous mushrooms and overlap in 1728 cases (45.24% of 3820 cases). Rule $R_1$ covers 3796 cases, thus rule $R_2$ adds only 3820-3796=24 new cases to $R_1$, which is 20.34% of 120 cases not covered by rule $R_1$. $R_6$ does not overlap with rule $R_1$ it adds 72 new cases to $R_1$, which is 60% of 120 cases not covered by rule $R_1$.

Thus, rules $R_1$&$R_2$, $R_1$&$R_6$ differ in an important characteristic. Heavy overlap of $R_1$ and $R_2$ (45.24%) *increases the confidence* in classification of those overlapped cases. The non-overlap of $R_1$ and $R_6$, *increases coverage* expanding the number of cases that the rule predicts two more than with adding $R_2$ to $R_1$.

For rules $R_2$-$R_7$, Table A3 shows overlap percentage for each pair of rules, the total number of distinct cases covered by two rules and the number of cases in their overlap. Only rules $R_2$ and $R_3$ heavily overlap. All other rules either do not overlap (7 pairs) or overlap in no more than 72 cases, which less than 10% of the total number of cases in each pair. As expected, many of the rules overlap with each other. The heaviest overlapping rules are $R_2$ and $R_3$, where the overlap is 64% together of the total cases predicted. This means that these two rules are closely related.

Table A4 shows the relations between rules $CR_1$-$CR_3$ from [29]. Rules CR1-CR3 use 5 attributes ($x_5$, $x_8$, $x_{12}$, $x_{20}$, $x_{21}$), while our rules R1-R7 use 7 attributes ($x_5$, $x_9$, $x_{15}$, $x_{19}$, $x_{20}$, $x_{21}$, $x_{22}$).

Table A3. Overlap percentage and number of cases for rules $R_2$-$R_7$.

|  | $R_3$ 1184 cases | $R_4$ 492 cases | $R_5$ 388 cases | $R_6$ 72 cases | $R_7$ 52 cases |
|---|---|---|---|---|---|
| $R_2$ 1752 cases | 64.00%<br>Total: 1784<br>Overlap: 1152 | 0%<br>Total: 2244<br>Overlap: 0 | 0.56%<br>Total: 2128<br>Overlap: 12 | 1.33%<br>Total: 1800<br>Overlap: 24 | 0%<br>Total: 1804<br>Overlap: 0 |
| $R_3$ 1184 cases |  | 0.96%<br>Total: 1660<br>Overlap: 16 | 0%<br>Total: 1572<br>Overlap: 0 | 0%<br>Total: 1256<br>Overlap: 0 | 0%<br>Total: 1236<br>Overlap: 0 |
| $R_4$ 492 cases |  |  | 8.01%<br>Total: 808<br>Overlap: 72 | 0%<br>Total: 564<br>Overlap: 0 | 8.8%<br>Total: 500<br>Overlap: 44 |
| $R_5$ 388 cases |  |  |  | 8.49%<br>Total: 424<br>Overlap: 36 | 0%<br>Total: 440<br>Overlap: 0 |
| $R_6$ 72 cases |  |  |  |  | 0%<br>Total: 124<br>Overlap: 0 |

Table A4. Relations between rules $CR_1$-$CR_3$.

|  |  | $CR_2$ 72 cases | $CR_3$ 912 cases |
|---|---|---|---|
| $CR_1$ 3796 cases | Overlap, % | 0% | 22.48% |
|  | Total cases | 3868 | 3844 |
|  | Overlap cases | 0 | 864 |
|  | Added cases | 72 | 48 |
|  | Added cases, % | 60% | 40% |
| $CR_2$ 72 cases | Overlap, % | 100% | 0% |
|  | Total cases | 72 | 984 |
|  | Overlap cases | 72 | 0 |
|  | Added cases | 0 | 912 |
|  | Added cases, % | 0% | 23.73% |

Rule $R_1$ is the same as rule $CR_1$ from [29]. Tables A2 and A3 show that rules $R_2$-$R_7$ are **more general** than rules $CR_2$ and $CR_3$ discovered in [29]. Rule $CR_2$ covers 72 cases and rule $CR_3$ covers 912 cases, while our rule $R_7$ with the smallest coverage cover 52 cases. Rules $R_2$-$R_7$ cover in total 2188 cases, while rules $CR_2$ and $CR_3$ cover only 984 cases together.

*A3. Experiment 3 with SRG2 algorithm on mushroom data with complimentary rules generation*

*A3.1. Rule generation*

Below we present rules generated at Step 3 first for the poisonous class and then complementary rules for the eatable class on Mushroom data with most frequent attributes generated from all 22 mushroom attributes. In this experiment the following attribute groups are used: Group A1: $x_9, x_5, x_7, x_{11}$; Group A2: $x_{13}, x_{14}, x_{15}, x_6$; Group A3: $x_1, x_2, x_4, x_{21}, x_{22}$.

According to the design of SRG algorithm it runs at the several levels of thresholds that can be set up by a user. We experimented with 3 level of precision: 75%. 85% and 95% with fixed level of coverage of 0.5%. This means that rules that have lower precision coverage are filtered out and not selected. For coverage with 3916 cases in the poisonous class it means that rules that cover less than 20 cases are filtered out considered as overfitting rules.

The thresholds 75%, 85%, and 95% limit only the low margin of the rule quality, but do not limit their upper level. Therefore, we also computed the actual precision and coverage reached for the poisonous class at each level.

Table A5 shows that rules at all levels missed only 4 cases from the poisonously class, giving 99.89% coverage and none of the poisonous cases was misclassified as eatable, giving 100% precision. All classification errors came from classifying some eatable cases as poisonous, that ranged from 800 cases for 75% threshold to 192 cases for 95% threshold.

Table A5. Results of sequential rule generation for poisonous class: Rules $R_1$-$R_{13}$.

| Characteristic | Level 1 | Level 2 | Level 3 |
|---|---|---|---|
| Low rule precision threshold. % | 75 | 85 | 95 |
| Low rule coverage threshold, % | 0.5 | 0.5 | 0.5 |
| Actual number of cases of poisonous class | 3916 | 3916 | 3916 |
| Number of rules selected | 14 | 14 | 13 |
| Number of cases covered by all rules | 4712 | 4520 | 4104 |
| Number of cases correctly classified by all rules | 3912 | 3912 | 3912 |
| Number of unclassified cases of the poisonous class | 4 | 4 | 4 |
| Number of misclassified cases by all rules | 800 | 608 | 192 |
| Actual coverage of the poisonous class, % | 99.89 | 99.89 | 99.89 |
| Actual precision, % | 100 | 100 | 100 |

We conducted the further analysis for 13 rules selected at level 3 with 95% threshold for rules. See Table A6. This table shows that all 192 misclassified cases belong to the rule $R_1$ that has 98.47% coverage and 95.2% precision. All other rules have 100% precision and coverage that smaller than for dominant rule $R_1$.

Table A6. Characteristics of discovered rules $R_1$-$R_{13}$ for poisonous class.

| Characteristic | $R_1$ | $R_2$ | $R_3$ | $R_4$ | $R_5$ | $R_6$ | $R_7$ |
|---|---|---|---|---|---|---|---|
| Precision, % | 95.02 | 100 | 100 | 100 | 100 | 100 | 100 |
| Coverage, % | 98.47 | 96.94 | 44.7 | 33.76 | 15.42 | 12.84 | 12.56 |
| Total cases predicted | 3856 | 3796 | 1752 | 1322 | 604 | 504 | 492 |
| Correct cases | 3664 | 3796 | 1752 | 1322 | 604 | 504 | 492 |
| Misclassified cases | 192 | 0 | 0 | 0 | 0 | 0 | 0 |
|  | $R_8$ | $R_9$ | $R_{10}$ | $R_{11}$ | $R_{12}$ | $R_{13}$ |  |
| Precision, % | 100 | 100 | 100 | 100 | 100 | 100 |  |
| Coverage, % | 10.01 | 9.91 | 1.48 | 1.23 | 1.12 | 0.92 |  |
| Total cases predicted | 392 | 388 | 58 | 48 | 44 | 36 |  |
| Correct cases | 392 | 388 | 58 | 48 | 44 | 36 |  |
| Misclassified cases | 0 | 0 | 0 | 0 | 0 | 0 |  |

**Rules $R_1$-$R_{13}$ (poisonous)**
**$R_1$:** $[(x_5 \neq 7)\ \&\ (x_7 \neq 2)\ \&\ (x_9 \neq 7)\ \&\ (x_{11} \neq 2)] \Rightarrow \mathbf{x} \in C_1$
**$R_2$:** $[(x_5=3) \vee (x_5=4) \vee (x_5=5) \vee (x_5=6) \vee (x_5=8) \vee (x_5=9)] \Rightarrow \mathbf{x} \in C_1$
**$R_3$:** $[(x_9=3) \vee (x_9=6)] \Rightarrow \mathbf{x} \in C_1$
**$R_4$:** $[(x_6 \neq 1)\ \&\ (x_{13} \neq 4)\ \&\ (x_{14} \neq 8)\ \&\ (x_{15} \neq 1)] \Rightarrow \mathbf{x} \in C_1$
**$R_5$:** $[(x_1 \neq 6)\ \&\ (x_2 \neq 3)\ \&\ (x_4 \neq 1)\ \&\ (x_{21} \neq 6)\ \&\ (x_{22}=7)] \Rightarrow \mathbf{x} \in C_1$
**$R_6$:** $[(x_9=5)\ \&\ (x_{11}=1)] \Rightarrow \mathbf{x} \in C_1$
**$R_7$:** $[(x_{15}=3) \vee (x_{15}=3) \vee (x_{15}=9)] \Rightarrow \mathbf{x} \in C_1$
**$R_8$:** $[(x_1=5)\ \&\ (x_4=2)\ \&\ (x_{21}=5)\ \&\ (x_{22} \neq 2)] \Rightarrow \mathbf{x} \in C_1$
**$R_9$:** $[(x_{21}=5)\ \&\ (x_{22}=1)] \Rightarrow \mathbf{x} \in C_1$
**$R_{10}$:** $[(x_6=3)\ \&\ (x_{13}=2)\ \&\ (x_{14} \neq 1)\ \&\ (x_{15} \neq 8)] \Rightarrow \mathbf{x} \in C_1$
**$R_{11}$:** $[(x_2=3)\ \&\ (x_{21}=2)\ \&\ (x_{22} \neq 6)] \Rightarrow \mathbf{x} \in C_1$
**$R_{12}$:** $[(x_1=1)\ \&\ (x_{21}=5)\ \&\ (x_{22} \neq 2)] \Rightarrow \mathbf{x} \in C_1$
**$R_{13}$:** $[(x_1 \neq 6)\ \&\ (x_2 \neq 1)\ \&\ (x_4 \neq 2)\ \&\ (x_{21}=5)\ \&\ (x_{22}=3)] \Rightarrow \mathbf{x} \in C_1$

The **rules R14 and R15** generated for the **eatable** class $C_2$ are presented below:
**$R_{14}$:** $[(x_5=1)\ \&\ (x_9 \neq 1)] \Rightarrow \mathbf{x} \in C_2$
**$R_{15}$:** $[(x_5=2)\ \&\ (x_9 \neq 1)] \Rightarrow \mathbf{x} \in C_2$

Table A7 shows the analysis of both R14 and R15 for class eatable class $C_2$. Rules $R_{14}$ and $R_{15}$ together cover and correctly predict all 192 cases misclassed by Rule $R_1$. While both $R_{14}$ and $R_{15}$ cover 336 cases each, these cases are different. In fact, rules $R_{14}$ and $R_{15}$ have 0% overlap and combined cover 672 cases of eatable class $C_2$.

Table A7. Characteristics of discovered rules $R_{14}$, $R_{15}$ for eatable class.

|  | $R_{14}$ | $R_{15}$ |
|---|---|---|
| Precision, % | 100 | 100 |

| Coverage, % | 7.98 | 7.98 |
|---|---|---|
| Total cases predicted | 336 | 336 |
| Correctly predicted cases | 366 | 336 |
| Misclassified cases | 0 | 0 |

*A3.2. Combining rules for two classes*

Now we have rules $R_1$-$R_{13}$ for the target class and rules $R_{14}$-$R_{15}$ for the non-target class and can accomplish step 3 of combining them. The only rule for the target class that misclassified some cases is $R_1$. So, we need to improve only this rule. It is done by creating a new rule $R_N$ that combines rule $R_1$ with $R_{14}$ and $R_{15}$ as follows

$$R_N(\mathbf{x}) = R_1(\mathbf{x})\ \&\ \neg\ (\ R_{14}(\mathbf{x}) \vee R_{15}(\mathbf{x})\ )$$

resulted in

$$R_N: [(x_5\neq 7)\ \&\ (x_7\neq 2)\ \&\ (x_9\neq 7)\ \&\ (x_{11}\neq 2)]\ \&$$
$$\neg\ (\ [(x_5=1)\ \&\ (x_9\neq 1)] \vee [(x_5=2)\ \&\ (x_9\neq 1)]\ ) => \mathbf{x} \in C_1$$

after putting actual rules $R_1$, $R_{14}$ and $R_{15}$ to the formula. Rule $R_N$ is false, $R_N(\mathbf{x})=0$, for all 192 cases $\mathbf{x}$ misclassified by rule $R_1$ as poisonous, for which $R_1(\mathbf{x})=1$, because for those $\mathbf{x}$ rule $R_{14}$ or $R_{15}$ is true. If each rule $R_{14}$ and $R_{15}$ would independently cover all 192 cases misclassed by $R_1$ then $R_N$ can be defined simpler in two ways by using any of these rules:

$$R_N(\mathbf{x}) = R_1(\mathbf{x})\ \&\ \neg\ R_{14}(\mathbf{x}),\quad R_N(\mathbf{x}) = R_1(\mathbf{x})\ \&\ \neg\ R_{15}(\mathbf{x}).$$

*A4. Experiment 4 with SRG3 on Mushroom data with all 30 random groups*

To evaluate this algorithm, we performed the test with 30 groups of 3 attributes were **randomly** generated from the total 22 mushroom attributes, as shown in Table A8. The algorithm generated all possible rules for these groups on the mushroom data. Then, with all these rules the rule combination process and selection process were ran and the final rules were selected, as shown in both Table A9 and Table A10. Below are the randomly generated groups and the result of this test.

**Rules R1-R7 (poisonous)**
**R₁:** $[(x_5=3) \vee (x_5=4) \vee (x_5=5) \vee (x_5=6) \vee (x_5=8) \vee (x_5=9)] \Rightarrow \mathbf{x} \in C_1$
**R₂:** $[(x_{12}=3)\ \&\ (x_{20}\neq 6)\ \&\ (x_7\neq 2)] \Rightarrow \mathbf{x} \in C_1$
**R₃:** $[(x_9=3) \vee (x_9=6)] \Rightarrow \mathbf{x} \in C_1$
**R₄:** $[(x_{19}\neq 6)\ \&\ (x_3=10)] \Rightarrow \mathbf{x} \in C_1$
**R₅:** $[(x_3=2)\ \&\ (x_{11}=1)] \Rightarrow \mathbf{x} \in C_1$
**R₆:** $[(x_{12}\neq 1)\ \&\ (x_{20}=5)\ \&\ (x_7=1)] \Rightarrow \mathbf{x} \in C_1$
**R₇:** $[(x_{16}=1)\ \&\ (x_{21}=2)\ \&\ (x_{11}\neq 7)] \Rightarrow \mathbf{x} \in C_1$

Table A8. 30 Randomly generated groups using 22 attributes.

| 16, 19, 5 | 16, 21, 11 | 1, 9, 10 | 19, 14, 3 | 3, 22, 1 | 14, 3, 11 |
|---|---|---|---|---|---|

| 22, 16, 14 | 22, 17, 20 | 1, 7, 19 | 13, 11, 5 | 5, 18, 14 | 16, 9, 18 |
|---|---|---|---|---|---|
| 7, 4, 22 | 6, 13, 17 | 12, 1, 14 | 4, 21, 14 | 2, 22, 1 | 6, 2, 8 |
| 13, 18, 19 | 6, 18, 8 | 5, 9, 8 | 16, 14, 21 | 1, 11, 3 | 9, 15, 6 |
| 6, 8, 9 | 7, 20, 19 | 22, 7, 5 | 9, 17, 3 | 12, 20, 7 | 4, 12, 15 |

Table A9. Sequential Rule Generation with 30 Randomly Generated Attribute Groupings of 3:

| Characteristic | Level 1 | Level 2 | Level 3 |
|---|---|---|---|
| Low rule precision threshold. % | 75 | 85 | 95 |
| Low rule coverage threshold, % | 0.5 | 0.5 | 0.5 |
| Actual number of cases of poisonous class | 3916 | 3916 | 3916 |
| Number of rules selected | 11 | 9 | 7 |
| Number of cases covered by all rules | 4716 | 4524 | 3916 |
| Number of cases correctly classified by all rules | 3916 | 3916 | 3916 |

At all levels tests there is no unclassified cases of the poisonous class, and misclassified cases by all rules, with **100% coverage** of the poisonous class and **100% accuracy**.

Table A10. Characteristics of discovered rules $R_1$-$R_7$ for poisonous class.

|  | $R_1$ | $R_2$ | $R_3$ | $R_4$ | $R_5$ | $R_6$ | $R_7$ |
|---|---|---|---|---|---|---|---|
| Precision, % | 100 | 100 | 100 | 100 | 100 | 100 | 100 |
| Coverage, % | 96.94 | 56.89 | 44.74 | 17.16 | 3.06 | 1.84 | 1.33 |
| Total cases predicted | 3796 | 2228 | 1752 | 672 | 120 | 72 | 52 |
| Correctly predicted cases | 3796 | 2228 | 1752 | 672 | 120 | 72 | 52 |
| Misclassified cases | 0 | 0 | 0 | 0 | 0 | 0 | 0 |

The analysis of the tables above shows that with 30 random groups the SRG algorithm was able to achieve **100% coverage** and **100% precision** requiring only 7 rules in the level 3 test. In addition, this result is also positive because it was able to pick up a better rule with small coverage (52 cases vs. 42 cases in the alternative rule).

*A5. Experiment 5 with SRG3 on Mushroom data with 13 most frequent attributes from 30 groups*

Here we tested the sequential rule generation algorithm SRG3 with the 13 most frequent attributes used in the 30 random triples test as seen above, where 7 rules reached 100% precision and 100% coverage. The 13 most frequent attributes were broken up into three different groups to ensure that the test would finish in reasonable time.

**Rules $R_1$-$R_7$ (poisonous)**:
**$R_1$:** $[(x_5=3) \vee (x_5=4) \vee (x_5=5) \vee (x_5=6) \vee (x_5=8) \vee (x_5=9)] \Rightarrow$ **x** $\in C_1$
**$R_2$:** $[(x_9=3) \vee (x_9=6)] \Rightarrow$ **x** $\in C_1$
**$R_3$:** $[(x_9=5) \& (x_{11}=1)] \Rightarrow$ **x** $\in C_1$
**$R_4$:** $[(x_{16}=1) \& (x_{19} \neq 2) \& (x_{20}=5) \& (x_{21}=5)] \Rightarrow$ **x** $\in C_1$
**$R_5$:** $[(x_{11}=2) \& (x_{12} \neq 4)] \Rightarrow$ **x** $\in C_1$
**$R_6$:** $[(x_{16}=1) \& (x_{19} \neq 2) \& (x_{20}=8) \& (x_{21}=2)] \Rightarrow$ **x** $\in C_1$
**$R_7$:** $[(x_3=10) \& (x_5=7)] \Rightarrow$ **x** $\in C_1$

Table A11. 13 used attributes in 30 random triple test results:

| Characteristic | Level 1 | Level 2 | Level 3 |
|---|---|---|---|

| Low rule precision threshold. % | 75 | 85 | 95 |
|---|---|---|---|
| Low rule coverage threshold, % | 0.5 | 0.5 | 0.5 |
| Actual number of cases of poisonous class | 3916 | 3916 | 3916 |
| Number of rules selected | 8 | 9 | 7 |
| Number of cases covered by all rules | 4700 | 4396 | 3900 |
| Number of cases correctly classified by all rules | 3900 | 3900 | 3900 |
| Number of unclassified cases of the poisonous class | 16 | 16 | 16 |
| Number of misclassified cases by all rules | 0 | 0 | 0 |
| Actual coverage of the poisonous class, % | 99.59 | 99.59 | 99.59 |
| Actual precision, % | 100 | 100 | 100 |

Table A12. Characteristics of discovered rules $R_1$-$R_7$ for poisonous class.

|  | $R_1$ | $R_2$ | $R_3$ | $R_4$ | $R_5$ | $R_6$ | $R_7$ |
|---|---|---|---|---|---|---|---|
| Precision, % | 100 | 100 | 100 | 100 | 100 | 100 | 100 |
| Coverage, % | 96.94 | 44.74 | 12.87 | 1.84 | 1.12 | 1.12 | 0.61 |
| Total cases predicted | 3796 | 1752 | 504 | 72 | 44 | 44 | 24 |
| Correctly predicted cases | 3796 | 1752 | 504 | 72 | 44 | 44 | 24 |
| Misclassified cases | 0 | 0 | 0 | 0 | 0 | 0 | 0 |

Here the SRG3 algorithm did not reduce the number of selected rules below 7.

*A6. Experiment 6 with SRG3 algorithm on mushroom data and 10-fold cross validation with generated rules*

Here we tested the abilities of SRG3 algorithm to generate beneficial rules in the 10-fold cross validation with the sequential triples attribute groups. These groups are G1: $\{x_1, x_2, x_3\}$; G2: $\{x_4, x_5, x_6\}$, …, G7: $\{x_{19}, x_{20}, x_{21}, x_{22}\}$. The 10-fold cross validation test was run with these groups to ensure that performance on the training data can be confirmed in the validation data. Table A13 shows the result of this test with 95% precision threshold for rule generation. In all 10 tests all rules provided 100% precision, 100% coverage of the target class.

Table A13. 10-fold cross validation results for sequential triple attribute groups

|  | Test 1 | Test 2 | Test 3 | Test 4 | Test 5 | Test 6 | Test 7 | Test 8 | Test 9 | Test 10 |
|---|---|---|---|---|---|---|---|---|---|---|
| Correctly predicted | 381 | 397 | 397 | 398 | 398 | 373 | 395 | 388 | 392 | 397 |
| Misclassified | 0 | 0 | 0 | 0 | 0 | 0 | 0 | 0 | 0 | 0 |
| Total classified | 381 | 397 | 397 | 398 | 398 | 373 | 395 | 388 | 392 | 397 |
| Validation cases | 812 | 812 | 812 | 812 | 812 | 812 | 813 | 813 | 813 | 813 |
| Rules for class $C_1$ | 4 | 4 | 4 | 4 | 4 | 4 | 4 | 4 | 4 | 4 |
| Rules for class $C_2$ | 0 | 0 | 0 | 0 | 0 | 0 | 0 | 0 | 0 | 0 |
| $C_1$ validation cases | 381 | 397 | 397 | 398 | 398 | 373 | 395 | 388 | 392 | 397 |

The table A13 shows that the 10-fold cross validation achieved **100% accuracy** in every test/fold. It generated and selected the **four rules** that were previously generated using all the data and the given attribute groups. This confirms the efficiency of the SRG algorithm to train and generate rules for newly added cases.

*A7. Experiment 7 with SRG3 algorithm on mushroom data and 10-fold cross validation with 30 random randomly generated triples*

Here we tested the abilities of SRG3 algorithm to generate beneficial rules using the 10-fold cross validation algorithm with 30 random triples of attribute groups as defined below. We ran this test four times to validate the accuracy of results. The result validation is necessary due to the variability of the randomly generated triples.

**Run 1**

Table A14. 30 Randomly generated triples of attributes for Run 1.

| 12, 15, 19 | 1, 4, 7 | 11, 15, 18 | 22, 3, 7 | 11, 14, 17 | 21, 3, 6 |
|---|---|---|---|---|---|
| 19, 22, 3 | 7, 11, 14 | 18, 21, 3 | 7, 10, 13 | 17, 21, 2 | 6, 9, 12 |
| 3, 6, 10 | 14, 17, 20 | 3, 6, 9 | 13, 16, 20 | 2, 5, 8 | 12, 16, 19 |
| 10, 13, 16 | 20, 2, 5 | 9, 12, 16 | 20, 1, 4 | 8, 12, 15 | 19, 22, 4 |
| 16, 20, 1 | 5, 8, 11 | 16, 19, 22 | 4, 8, 11 | 15, 18, 21 | 4, 7, 10 |

Table A15. 10-fold cross validation results for sequential triple attribute groups (95% precision threshold)

|  | Test 1 | Test 2 | Test 3 | Test 4 | Test 5 | Test 6 | Test 7 | Test 8 | Test 9 | Test 10 |
|---|---|---|---|---|---|---|---|---|---|---|
| Correctly predicted | 393 | 394 | 409 | 386 | 385 | 394 | 383 | 391 | 385 | 396 |
| Misclassified | 0 | 0 | 0 | 0 | 0 | 0 | 0 | 0 | 0 | 0 |
| Total predicted | 393 | 394 | 409 | 386 | 385 | 394 | 383 | 391 | 385 | 396 |
| Rules selected for C1 | 5 | 5 | 5 | 5 | 5 | 5 | 5 | 5 | 5 | 5 |
| Rules selected for C2 | 0 | 0 | 0 | 0 | 0 | 0 | 0 | 0 | 0 | 0 |
| Target validation cases | 393 | 394 | 409 | 386 | 385 | 394 | 383 | 391 | 385 | 396 |

In all 10 tests all rules provided 100% precision, 100% coverage of the target class and 100% accuracy.

*Rules generated* using 10-Fold Cross Validation:
**CR$_1$ = R$_1$:** $[(x_5=3) \lor (x_5=4) \lor (x_5=5) \lor (x_5=6) \lor (x_5=8) \lor (x_5=9)] \Rightarrow \mathbf{x} \in C_1$
**R$_2$:** $[(x_4 \neq 2) \,\&\, (x_{20} = 5)] \Rightarrow \mathbf{x} \in C_1$
**R$_3$:** $[(x_{12} = 3) \,\&\, (x_{16} = 1) \,\&\, (x_{19} \neq 6)] \Rightarrow \mathbf{x} \in C_1$
**R$_4$:** $[(x_4 \neq 2) \,\&\, (x_7 = 2) \,\&\, (x_{10} = 1)] \Rightarrow \mathbf{x} \in C_1$
**R$_5$:** $[(x_{19} \neq 6) \,\&\, (x_3 = 10)] \Rightarrow \mathbf{x} \in C_1$
Complexity 16/3520 = 0.004545

**Run 2**

Table A16. 30 Randomly Generated Triples for Run 2.

| 11, 14, 17 | 21, 3, 6 | 10, 13, 16 | 21, 2, 5 | 9, 13, 16 | 20, 1, 4 |
|---|---|---|---|---|---|
| 17, 20, 2 | 6, 9, 12 | 16, 20, 1 | 5, 8, 12 | 16, 19, 22 | 4, 8, 11 |
| 2, 5, 8 | 12, 16, 19 | 1, 4, 8 | 12, 15, 18 | 22, 4, 7 | 11, 14, 18 |
| 8, 12, 15 | 19, 22, 3 | 8, 11, 14 | 18, 21, 3 | 7, 10, 13 | 18, 21, 2 |
| 15, 18, 21 | 3, 7, 10 | 14, 17, 21 | 3, 6, 9 | 13, 17, 20 | 2, 5, 9 |

Table A17. 10-fold cross validation results for sequential triple attribute groups (95% precision threshold).

| Characteristic | Test 1 | Test 2 | Test 3 | Test 4 | Test 5 | Test 6 | Test 7 | Test 8 | Test 9 | Test 10 |
|---|---|---|---|---|---|---|---|---|---|---|
| Correctly predicted | 395 | 405 | 380 | 388 | 400 | 396 | 375 | 396 | 398 | 383 |
| Incorrectly predicted | 0 | 0 | 0 | 0 | 0 | 0 | 0 | 0 | 0 | 0 |
| Total predicted | 395 | 405 | 380 | 388 | 400 | 396 | 375 | 396 | 398 | 383 |
| Rules selected for C1 | 5 | 5 | 5 | 5 | 5 | 5 | 5 | 5 | 5 | 5 |
| Rules selected for C2 | 0 | 0 | 0 | 0 | 0 | 0 | 0 | 0 | 0 | 0 |
| Validation case of class 1 | 395 | 405 | 380 | 388 | 400 | 396 | 375 | 396 | 398 | 383 |

At all 10 tests precision of all rules is 100% and target class coverage by rules is 100%.

Rules generated using 10-Fold Cross Validation:
**CR$_1$ = R$_1$:** $[(x_5=3) \vee (x_5=4) \vee (x_5=5) \vee (x_5=6) \vee (x_5=8) \vee (x_5=9)] \Rightarrow \mathbf{x} \in C_1$
**R$_2$:** $[(x_4 \neq 2) \& (x_{20} = 5)] \Rightarrow \mathbf{x} \in C_1$
**R$_3$:** $[(x_{12} = 3) \& (x_{18} \neq 3)] \Rightarrow \mathbf{x} \in C_1$
**R$_4$:** $[(x_8 \neq 1) \& (x_{14} \neq 8)] \Rightarrow \mathbf{x} \in C_1$
**R$_5$:** $[(x_{22} = 2) \& (x_4 \neq 2)] \Rightarrow \mathbf{x} \in C_1$
Complexity = 14/3533 = 0.00396

**Run 3**

Table A18. 30 Randomly Generated Triples for Run 3.

| 17, 21, 2 | 6, 9, 13 | 17, 20, 1 | 5, 9, 12 | 16, 19, 1 | 5, 8, 11 |
|---|---|---|---|---|---|
| 2, 5, 8 | 13, 16, 19 | 1, 5, 8 | 12, 15, 18 | 1, 4, 7 | 11, 14, 18 |
| 8, 12, 15 | 19, 22, 4 | 8, 11, 14 | 18, 22, 3 | 7, 10, 14 | 18, 21, 2 |
| 15, 18, 22 | 4, 7, 10 | 14, 18, 21 | 3, 6, 9 | 14, 17, 20 | 2, 6, 9 |
| 22, 3, 6 | 10, 13, 17 | 21, 2, 5 | 9, 13, 16 | 20, 1, 5 | 9, 12, 15 |

Table A19. 10-fold cross validation results for sequential triple attribute groups (95% precision threshold).

| Characteristic | Test 1 | Test 2 | Test 3 | Test 4 | Test 5 | Test 6 | Test 7 | Test 8 | Test 9 | Test 10 |
|---|---|---|---|---|---|---|---|---|---|---|
| Correctly predicted | 378 | 398 | 394 | 406 | 397 | 397 | 383 | 382 | 390 | 391 |
| Misclassified | 0 | 0 | 0 | 0 | 0 | 0 | 0 | 0 | 0 | 0 |
| Total predicted | 378 | 398 | 394 | 406 | 397 | 397 | 383 | 382 | 390 | 391 |
| Rules selected for C1 | 4 | 4 | 4 | 4 | 4 | 4 | 4 | 4 | 4 | 4 |
| Rules selected for C2 | 0 | 0 | 0 | 0 | 0 | 0 | 0 | 0 | 0 | 0 |
| Target validation cases | 378 | 398 | 394 | 406 | 397 | 397 | 383 | 382 | 390 | 391 |

In all 10 tests all rules provided 100% precision, 100% coverage of the target class.

Rules generated using 10-Fold Cross Validation:
**CR$_1$ = R$_1$:** $[(x_5=3) \vee (x_5=4) \vee (x_5=5) \vee (x_5=6) \vee (x_5=8) \vee (x_5=9)] \Rightarrow \mathbf{x} \in C_1$
**R$_2$:** $[(x_5 \neq 2) \& (x_{20} = 5)] \Rightarrow \mathbf{x} \in C_1$
**R$_3$:** $[(x_{13} = 2) \& (x_{16} = 1) \& (x_{19} \neq 6)] \Rightarrow \mathbf{x} \in C_1$
**R$_4$:** $[(x_{15} = 8) \& (x_{22} = 2)] \Rightarrow \mathbf{x} \in C_1$
Complexity = 13/3525 = 0.003688

## Run 4

Table A20. 30 Randomly Generated Triples for Run 4.

| 3, 6, 9    | 14, 17, 20 | 2, 5, 9    | 13, 16, 19 | 2, 5, 8    | 12, 15, 19 |
|------------|------------|------------|------------|------------|------------|
| 9, 13, 16  | 20, 1, 5   | 9, 12, 15  | 19, 1, 4   | 8, 11, 15  | 19, 22, 3  |
| 16, 19, 22 | 5, 8, 11   | 15, 19, 22 | 4, 7, 10   | 15, 18, 21 | 3, 6, 10   |
| 22, 4, 7   | 11, 14, 18 | 22, 3, 6   | 10, 14, 17 | 21, 2, 6   | 10, 13, 16 |
| 7, 10, 14  | 18, 21, 2  | 6, 10, 13  | 17, 20, 2  | 6, 9, 12   | 16, 20, 1  |

Table A21. 10-fold cross validation results for sequential triple attribute groups (95% precision threshold).

| Characteristic | Test 1 | Test 2 | Test 3 | Test 4 | Test 5 | Test 6 | Test 7 | Test 8 | Test 9 | Test 10 |
|---|---|---|---|---|---|---|---|---|---|---|
| Correctly predicted | 398 | 398 | 374 | 379 | 401 | 388 | 401 | 408 | 384 | 385 |
| Misclassified | 0 | 0 | 0 | 0 | 0 | 0 | 0 | 0 | 0 | 0 |
| Total predicted | 398 | 398 | 374 | 379 | 401 | 388 | 401 | 408 | 384 | 385 |
| Rules selected for Class 1 | 4 | 4 | 4 | 4 | 4 | 4 | 4 | 4 | 4 | 4 |
| Rules selected for Class 2 | 0 | 0 | 0 | 0 | 0 | 0 | 0 | 0 | 0 | 0 |
| Validation cases of class 1 | 398 | 398 | 374 | 379 | 401 | 388 | 401 | 408 | 384 | 385 |

In all 10 tests all rules provided 100% precision and 100% coverage of the target class.

Rules generated using 10-Fold Cross Validation:
**CR$_1$ = R$_1$:** $[(x_5=3) \lor (x_5=4) \lor (x_5=5) \lor (x_5=6) \lor (x_5=8) \lor (x_5=9)] \Rightarrow \mathbf{x} \in C_1$
**R$_2$:** $[(x_1 \neq 6)$ & $(x_{16} = 1)$ & $(x_{20} = 5)] \Rightarrow \mathbf{x} \in C_1$
**R$_3$:** $[(x_{13} = 2)$ & $(x_{16} = 1)$ & $(x_{19} \neq 6)] \Rightarrow \mathbf{x} \in C_1$
**R$_4$:** $[(x_{15} = 8)$ & $(x_{22} = 2)] \Rightarrow \mathbf{x} \in C_1$
Complexity = 14/3531 = 0.00396

Table A22. Summary of runs.

| Characteristic | Run 1 | Run 2 | Run 3 | Run 4 |
|---|---|---|---|---|
| Number of Rules | 4 | 5 | 4 | 4 |
| Number of clauses used in rules | 16 | 14 | 13 | 14 |
| Cases covered by rules in 10-fold CV | 3520 | 3533 | 3525 | 3531 |
| Incorrectly predicted cases | 0 | 0 | 0 | 0 |
| Complexity of rules | 16/3520 = 0.0045 | 14/3533 = 0.0037 | **13/3525 = 0.0037** | 14/3531 = 0.00396 |

In all runs all rules provided 100% precision, 100% coverage of the target class and 100% accuracy. While the tests are accurate and precise, the generated rules have a moderate amount of variation in complexity and attributes used. The variation is due to the random group generation and the 10-fold data partition.

### A.7.1. Rule generation

Below we present rules generated at Step 3 first for the poisonous class and then complementary rules for the eatable class on Mushroom data with most frequent attributes generated from all 22 mushroom attributes. In this experiment the following attribute groups are used: Group A1: $x_9, x_5, x_7, x_{11}$; Group A2: $x_{13}, x_{14}, x_{15}, x_6$; Group A3: $x_1, x_2, x_4, x_{21}, x_{22}$.

According to the design of SRG algorithm it runs at the several levels of thresholds that can be set up by a user. We experimented with 3 level of precision: 75%. 85% and 95% with fixed level of coverage of 0.5%. This means that rules that have lower precision coverage are filtered out and not selected. For coverage with 3916 cases in the poisonous class it means that rules that cover less than 20 cases are filtered out considered as overfitting rules.

The thresholds 75%, 85%, and 95% limit the low margin of the rule quality, but not their upper level. Therefore, we computed the actual precision and coverage reached for the poisonous class at each level. Table A23 shows that rules at all levels missed only 4 cases from the poisonously class, giving 99.89% coverage and none of the poisonous cases was misclassified, giving 100% precision. All misclassified are eatable cases that ranged from 800 cases for 75% threshold to 192 cases for 95% threshold.

Table A23. Results of sequential rule generation for poisonous class: Rules $R_1$-$R_{13}$.

| Characteristic | Level 1 | Level 2 | Level 3 |
|---|---|---|---|
| Low rule precision threshold. % | 75 | 85 | 95 |
| Low rule coverage threshold, % | 0.5 | 0.5 | 0.5 |
| Actual number of cases of poisonous class | 3916 | 3916 | 3916 |
| Number of rules selected | 14 | 14 | 13 |
| Number of cases covered by all rules | 4712 | 4520 | 4104 |
| Number of cases correctly classified by all rules | 3912 | 3912 | 3912 |
| Number of unclassified cases of the poisonous class | 4 | 4 | 4 |
| Number of misclassified cases by all rules | 800 | 608 | 192 |
| Actual coverage of the poisonous class, % | 99.89 | 99.89 | 99.89 |
| Actual precision, % | 100 | 100 | 100 |

We conducted the further analysis for 13 rules selected at level 3 with 95% threshold for rules. See Table A24. This table shows that all 192 misclassified cases belong to the rule $R_1$ that has 98.47% coverage and 95.2% precision. All other rules have 100% precision and coverage that smaller than for dominant rule $R_1$.

Table A24. Characteristics of discovered rules $R_1$-$R_{13}$ for poisonous class.

| Characteristic | $R_1$ | $R_2$ | $R_3$ | $R_4$ | $R_5$ | $R_6$ | $R_7$ |
|---|---|---|---|---|---|---|---|
| Precision, % | 95.02 | 100 | 100 | 100 | 100 | 100 | 100 |
| Coverage, % | 98.47 | 96.94 | 44.7 | 33.76 | 15.42 | 12.84 | 12.56 |
| Total cases predicted | 3856 | 3796 | 1752 | 1322 | 604 | 504 | 492 |
| Correct cases | 3664 | 3796 | 1752 | 1322 | 604 | 504 | 492 |
| Misclassified cases | 192 | 0 | 0 | 0 | 0 | 0 | 0 |
| | $R_8$ | $R_9$ | $R_{10}$ | $R_{11}$ | $R_{12}$ | $R_{13}$ | |
| Precision, % | 100 | 100 | 100 | 100 | 100 | 100 | |
| Coverage, % | 10.01 | 9.91 | 1.48 | 1.23 | 1.12 | 0.92 | |
| Total cases predicted | 392 | 388 | 58 | 48 | 44 | 36 | |
| Correct cases | 392 | 388 | 58 | 48 | 44 | 36 | |
| Misclassified cases | 0 | 0 | 0 | 0 | 0 | 0 | |

**Rules R$_1$-R$_{13}$ (poisonous)**

**R$_1$:** [($x_5 \neq 7$) & ($x_7 \neq 2$) & ($x_9 \neq 7$) & ($x_{11} \neq 2$)] $\Rightarrow$ **x** $\in$ C$_1$
**R$_2$:** [($x_5$=3) $\vee$ ($x_5$=4) $\vee$ ($x_5$=5) $\vee$ ($x_5$=6) $\vee$ ($x_5$=8) V ($x_5$=9)] $\Rightarrow$ **x** $\in$ C$_1$
**R$_3$:** [($x_9$=3) $\vee$ ($x_9$= 6)] $\Rightarrow$ **x** $\in$ C$_1$
**R$_4$:** [($x_6 \neq 1$) & ($x_{13} \neq 4$) & ($x_{14} \neq 8$) & ($x_{15} \neq 1$)] $\Rightarrow$ **x** $\in$ C$_1$
**R$_5$:** [($x_1 \neq 6$) & ($x_2 \neq 3$) & ($x_4 \neq 1$) & ($x_{21} \neq 6$) & ($x_{22}$=7)] $\Rightarrow$ **x** $\in$ C$_1$
**R$_6$:** [($x_9$=5) & ($x_{11}$=1)] $\Rightarrow$ **x** $\in$ C$_1$
**R$_7$:** [($x_{15}$=3) $\vee$ ($x_{15}$=3) $\vee$ ($x_{15}$=9)] $\Rightarrow$ **x** $\in$ C$_1$
**R$_8$:** [($x_1$=5) & ($x_4$=2) & ($x_{21}$=5) & ($x_{22} \neq 2$)] $\Rightarrow$ **x** $\in$ C$_1$
**R$_9$:** [($x_{21}$=5) & ($x_{22}$=1)] $\Rightarrow$ **x** $\in$ C$_1$
**R$_{10}$:** [($x_6$=3) & ($x_{13}$=2) & ($x_{14} \neq 1$) & ($x_{15} \neq 8$)] $\Rightarrow$ **x** $\in$ C$_1$
**R$_{11}$:** [($x_2$=3) & ($x_{21}$=2) & ($x_{22} \neq 6$)] $\Rightarrow$ **x** $\in$ C$_1$
**R$_{12}$:** [($x_1$=1) & ($x_{21}$=5) & ($x_{22} \neq 2$)] $\Rightarrow$ **x** $\in$ C$_1$
**R$_{13}$:** [($x_1 \neq 6$) & ($x_2 \neq 1$) & ($x_4 \neq 2$) & ($x_{21}$=5) & ($x_{22}$=3)] $\Rightarrow$ **x** $\in$ C$_1$

The **rules R14 and R15** generated for the **eatable** class C$_2$ are presented below:

**R$_{14}$:** [($x_5$=1) & ($x_9 \neq 1$)] => **x** $\in$ C$_2$
**R$_{15}$:** [($x_5$=2) & ($x_9 \neq 1$)] => **x** $\in$ C$_2$

Table A25 shows the analysis of both R14 and R15 for class eatable class C$_2$. Rules R$_{14}$ and R$_{15}$ together cover and correctly predict all 192 cases misclassed by Rule R$_1$. While both R$_{14}$ and R$_{15}$ cover 336 cases each, these cases are different. In fact, rules R$_{14}$ and R$_{15}$ have 0% overlap and combined cover 672 cases of eatable class C$_2$.

Table A25. Characteristics of discovered rules R$_{14}$, R$_{15}$ for eatable class.

|  | R$_{14}$ | R$_{15}$ |
| --- | --- | --- |
| Precision, % | 100 | 100 |
| Coverage, % | 7.98 | 7.98 |
| Total cases predicted | 336 | 336 |
| Correctly predicted cases | 366 | 336 |
| Misclassified cases | 0 | 0 |

*A.7.2. Combining rules for two classes*

Now we have rules R$_1$-R$_{13}$ for the target class and rules R$_{14}$-R$_{15}$ for the non-target class and can accomplish step 3 of combining them. The only rule for the target class that misclassified some cases is R$_1$. So, we need to improve only this rule. It is done by creating a new rule R$_N$ that combines rule R$_1$ with R$_{14}$ and R$_{15}$ as follows

$$R_N(\mathbf{x}) = R_1(\mathbf{x}) \;\&\; \neg \;( R_{14}(\mathbf{x}) \vee R_{15}(\mathbf{x}) )$$

resulted in

$$R_N: [(x_5 \neq 7) \& (x_7 \neq 2) \& (x_9 \neq 7) \& (x_{11} \neq 2)] \&$$
$$\neg ( [(x_5=1) \& (x_9 \neq 1)] \vee [(x_5=2) \& (x_9 \neq 1)] ) \Rightarrow \mathbf{x} \in C_1$$

after putting actual rules $R_1$, $R_{14}$ and $R_{15}$ to the formula.

Rule $R_N$ is false, $R_N(\mathbf{x})=0$, for all 192 cases $\mathbf{x}$ misclassified by rule $R_1$ as poisonous, for which $R_1(\mathbf{x})=1$, because for those $\mathbf{x}$ rule $R_{14}$ or $R_{15}$ is true. If each rule $R_{14}$ and $R_{15}$ would independently cover all 192 cases misclassed by $R_1$ then $R_N$ can be defined simpler in two ways by using any of these rules:

$$R_N(\mathbf{x}) = R_1(\mathbf{x}) \& \neg R_{14}(\mathbf{x}), \quad R_N(\mathbf{x}) = R_1(\mathbf{x}) \& \neg R_{15}(\mathbf{x}).$$

*A8. Experiment 8 with algorithm SRG4 based on expert selected groups*

The results of this test are shown in tables A26 and A27. The groups used are: Group 1 (Cap): $\{x_1, x_2, x_3\}$; Group 2 (Odor): $\{x_5\}$; Group 3 (Gill): $\{x_6, x_7, x_8, x_9\}$; Group 4 (Stalk): $\{x_{10}, x_{11}, x_{13}, x_{15}\}$; Group 5 (Veil): $\{x_{17}, x_{18}, x_{19}\}$; Group 6 (Spore): $\{x_{20}\}$; Group 7 (Distribution): $\{x_{21}, x_{22}\}$. The analysis of tables A26 and A27 shows that using the groups generated by the biologist the SRG algorithm was able to get a good result of 99.81% coverage of the target class with 100% precision using just **3 rules**. Although this is considered to be a good result, it did not cover the whole of class $C_1$. One way this can be achieved is by altering the coverage threshold to a value less than 0.5%. Making this adjustment would allow more small coverage rules to be generated an in turn would allow the combination phase to combine these smaller coverage rules with larger coverage rules to produce high coverage and general rules that cover more class $C_1$ cases.

Table A26. Sequential rule generation test using expert groups from Biologist (95% precision threshold)

|  | Level 1 | Level 2 | Level 3 |
|---|---|---|---|
| Low rule precision threshold. % | 75 | 85 | 95 |
| Low rule coverage threshold, % | 0.5 | 0.5 | 0.5 |
| Actual number of cases of poisonous class | 3916 | 3916 | 3916 |
| Number of rules selected | 3 | 4 | 3 |
| Number of cases covered by all rules | 3908 | 3908 | 3908 |
| Number of cases correctly classified by all rules | 3908 | 3908 | 3908 |
| Number of unclassified cases of the poisonous class | 8 | 8 | 8 |
| Number of misclassified cases by all rules | 99.81 | 99.81 | 99.81 |
| Actual coverage of the poisonous class, % | 100 | 100 | 100 |
| Actual precision, % | 100 | 100 | 100 |

Table A27. Characteristics of discovered rules $R_1$-$R_3$ for poisonous class (95% precision threshold).

| Characteristic | R1 | R2 | R3 |
|---|---|---|---|
| Precision, % | 100 | 100 | 100 |
| Coverage, % | 96.94 | 1.84 | 12.97 |
| Total cases predicted | 3796 | 72 | 508 |
| Correctly predicted cases | 3796 | 72 | 508 |
| Misclassified cases | 0 | 0 | 0 |

**R₁:** $[(x_5=3) \lor (x_5=4) \lor (x_5=5) \lor (x_5=6) \lor (x_5=8) \lor (x_5=9)] \Rightarrow \mathbf{x} \in C_1$
**R₂:** $[(x_{20}=5)] \Rightarrow \mathbf{x} \in C_1$
**R₃:** $[(x_{11} \neq 1) \& (x_{13} \neq 4) \& (x_{15} \neq 8)] \Rightarrow \mathbf{x} \in C_1$

### A9. Experiment 9 with SRG5 and 7 successful attributes

Here we tested SRG algorithm with the 7 used attributes in our previous 100% precision test. This test was run with these attributes in hopes that only using the 7 used attributes will allow the rule generation process to generate and select less rules while keeping the 100% precision and total coverage of the target class.

The 7 attributes were broken up into two different groups to be able to finish the test in reasonable time, where group 1: $x_5$, $x_9$, $x_{15}$; and Group 2: $x_{19}$, $x_{20}$, $x_{21}$, $x_{22}$. The results are shown in Tables A28 and A29.

**Rules R₁-R₇ (poisonous):**
**R₁:** $[(x_5=3) \lor (x_5=4) \lor (x_5=5) \lor (x_5=6) \lor (x_5=8) \lor (x_5=9)] \Rightarrow \mathbf{x} \in C_1$
**R₂:** $[(x_9=3) \lor (x_9=6)] \Rightarrow \mathbf{x} \in C_1$
**R₃:** $[(x_{19}=2) \& (x_{20}=8) \& (x_{21} \neq !2) \& (x_{22} \neq !2)] \Rightarrow \mathbf{x} \in C_1$
**R₄:** $[(x_{15}=3) \lor (x_{15}=2) \lor (x_{15}=9)] \Rightarrow \mathbf{x} \in C_1$
**R₅:** $[(x_{19} \neq 2) \& (x_{20} \neq 6) \& (x_{21}=5) \& (x_{22}=1)] \Rightarrow \mathbf{x} \in C_1$
**R₆:** $[(x_{19}=6) \& (x_{20}=5) \& (x_{21} \neq 1) \& (x_{22} \neq 6)] \Rightarrow \mathbf{x} \in C_1$
**R₇:** $[(x_{20}=8) \& (x_{21}=2) \& (x_{22} \neq 6)] \Rightarrow \mathbf{x} \in C_1$

Table A28. Sequential rule generation test using successful attributes.

|  | Level 1 | Level 2 | Level 3 |
|---|---|---|---|
| Low rule precision threshold. % | 75 | 85 | 95 |
| Low rule coverage threshold, % | 0.5 | 0.5 | 0.5 |
| Actual number of cases of poisonous class | 3916 | 3916 | 3916 |
| Number of rules selected | 10 | 11 | 7 |
| Number of cases covered by all rules | 4716 | 4492 | 3916 |
| Number of cases correctly classified by all rules | 3916 | 3916 | 3916 |
| Number of unclassified cases of the poisonous class | 0 | 0 | 0 |

At all levels all rules provided **100% precision, 100% coverage** of the target class.

Table A29. Characteristics of discovered rules R₁-R₇ for poisonous class.

|  | R1 | R2 | R3 | R4 | R5 | R6 | R7 |
|---|---|---|---|---|---|---|---|
| Precision, % | 100 | 100 | 100 | 100 | 100 | 100 | 100 |
| Coverage, % | 96.94 | 44.74 | 30.23 | 12.56 | 9.91 | 1.84 | 1.33 |
| Total cases predicted | 3796 | 1752 | 1184 | 492 | 388 | 72 | 52 |
| Correctly predicted cases | 3796 | 1752 | 1184 | 492 | 388 | 72 | 52 |
| Misclassified cases | 0 | 0 | 0 | 0 | 0 | 0 | 0 |

These tables allow to conclude that the sequential rule generation algorithm using the attribute groups created by using the 7 attributes was unable to reduce the number of rules selected to cover all of class $C_1$. This result is likely due to the SRG2 process used. Other existing or new versions SRG algorithm can be more successful in future.

*A10. Experiment 10 with SPG 5 and comparison of rule complexity*

In this experiment we used SRG5 algorithm based on SRG1 algorithm, with attributes that have been successful for mushroom data in [29,32]. We compared the complexity of rules generated in this process with rules from [29,32], which we denote as **CR rules**.

The formulas for computing complexity of rules and sets of rules are given in section 5.2. Several sets of rules were generated in [29,32]. We use only the final rules from [29,32] listed below in our notation.

**CR Rules** [29,32]
**CR$_1$:** $[(x_5=3) \vee (x_5=4) \vee (x_5=5) \vee (x_5=6) \vee (x_5=8) \vee (x_5=9)] \Rightarrow \mathbf{x} \in C_1$,
Complexity $6/3796 = 0.0016$
**CR$_2$:** $[(x_{20}=5)] \Rightarrow \mathbf{x} \in C_1$, Complexity $1/72 = 0.014$
**CR$_3$:** $[(x_8=2) \& (x_{12}=3)] \vee [(x_8=2) \& (x_{12}=2)] \vee [(x_8=2) \& (x_{21}=2)] \Rightarrow \mathbf{x} \in C_1$
Complexity $6/912 = 0.0066$
Complexity of the set of rules: $(6+1+6)/(3796+72+912) = 13/4780 = \mathbf{0.0027}$

Attribute Groups that we used in SRG algorithm are G1 = $\{x_5\}$; G2 = $\{x_{20}\}$; G3 = $\{x_8, x_{12}, x_{21}\}$, which directly correspond to CR rules above.

**Our Rules**
**R$_1$:** $[(x_5=3) \vee (x_5=4) \vee (x_5=5) \vee (x_5=6) \vee (x_5=8) \vee (x_5=9)] \Rightarrow \mathbf{x} \in C_1$
Complexity $6/3796 = 0.0016$
**R$_2$:** $[(x_{20}=5)] \Rightarrow \mathbf{x} \in C_1$, Complexity $1/72 = 0.014$
**R$_3$:** $[(x_{12}=3) \& (x_{21}=5)] \Rightarrow \mathbf{x} \in C_1$, Complexity $2/1544 = 0.0013$
**R$_4$:** $[(x_8 \neq 1) \& (x_{21}=2)] \Rightarrow \mathbf{x} \in C_1$, Complexity $2/16 = 0.125$

Complexity of a set of rules = $(6+1+2+2)/(3796+72+1544+16) = 11/5428 = \mathbf{0.002}$

The algorithm SRG1 generated **simpler rules** (11 clauses vs. 13 clauses) with **the same precision** as in in [29,32] using the attribute groups derived from CR rules. This result shows that the algorithm SRG1 can generate rules that are **less complex** than CR rules and suggests that better rules are possible with more testing and preprocessing of attribute groups.

*A11. Toolkit*

The toolkit includes the **Data Type Editor** integrated with visualization system **VisCanvas** 2.0 [2,3] for multidimensional data visualization based on the adjustable parallel coordinates. The data type editor supports saving data in the explainable measurement coding format for pattern discovery and data visualization. Fig. A30 illustrates setting up and applying a coding scheme for data that converts letter grades for 4 classes $X_1$-$X_4$ as follows: A to 4, B to 3, C to 2, and so on.

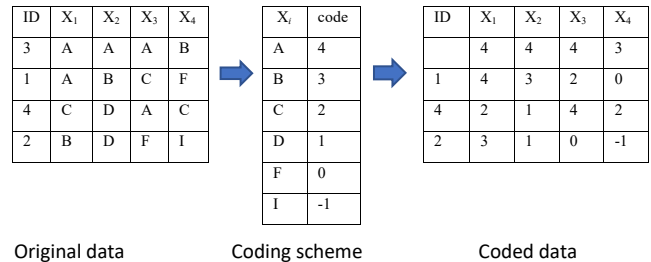

    Original data        Coding scheme        Coded data

Fig. A30. Example of applying coding schema for class grades.

The toolkit supports Nominal, Ordinal, Interval, and Ratio data measurement types. The attributes of the absolute measurement data type are encoded as the ratio data type because it differs from it only by presence of the fixed measurement unit.

The data type editor provides helpful descriptions and examples in case the user is not familiar with these data types. The user interface allows a user: to assign data type for each attribute, and to group values of each attribute. The **scheme loader** allows a user to assign data measurement type: nominal, ordinal, interval, and ratio.

A typical example of **mixed data** are the mushroom data [9], which contain 8124 instances and 22 attributes. These attributes include **nominal** data such as habitat (grass, leaves, meadows, paths, urban, waste, woods), **ordinal** data, such as gill size (broad, narrow) and gill spacing (crowded, close, distant), **absolute** data such as the number of rings (0,1,2), which the scheme loader treats as **ratio** data.

The colors of different parts of the mushroom such as cap color represent an interesting data type. It can be treated as: (1) nominal (red, blue, green and so on), (2) three numeric attributes like R,G,B, (3) some scalar function from R,G,B like R+G+B, and (4) a single numeric **ratio** data type that uses wavelength. The first one does not capture the similarity between colors, the second one is expanding the number of attributes, the third one corrupts similarity relations between some colors, and the last one is the most physically meaningful.

Since, each color covers a wavelength interval, grouping wavelength values according to colors is a natural way to encode the colors. These groups are ordered and can be encoded by integers starting from 0. In general, grouping attribute values decreases the space size and run time of the algorithms.

Manual coding is time consuming and tedious work for the tasks with many attributes and multiple values of each attribute. The toolkit allows to speed up this process. The editor has the "All Ordinal" and "All Nominal" options that allow to assign initially

Nominal or Ordinal type and to assign integer code values from 1 to *n* to all attributes with abilities to edit this assignment later.

**Grouping**. Fig. A31 illustrates grouping and binary coding keys for a **nominal** attribute. Fig. A32 illustrates setting up groups for the numeric interval and ratio attributes by creating **intervals** where user sets on the starting value for the group and the length of its interval. Original values of the attributes may not correctly represent its data type for the task. The user can select keeping the existing values or generate new ones.

**Hierarchy of attribute groups**. When we have hundreds of attributes, a hierarchy of attributes allows to deal with them efficiently. The system supports a user in constructing a hierarchy and picking up a level at which attributes will be visualized.

| Nominal Grouping | | | Scheme Loader (Attribute: X1) | |
|---|---|---|---|---|
| | Key | Group | Key | Value |
| | s | 1 | x | 01b |
| | y | 1 | b | 10b |
| | f | 2 | s | 01b |
| | g | 2 | f | 01b |

Fig. A31. Grouping keys and assigning binary codes for a nominal attribute.

| Interval Group Starting | | | Scheme Loader | |
|---|---|---|---|---|
| Group Number | Starting Point | Interval To First | Key | Value |
| 1 | 104 | 8 | 1 | 243 |
| | | | 2 | 245 |
| | | | 5 | 249 |
| | | | 8 | 255 |
| | | | 4 | 263 |

(a) Setting up intervals scheme.         (b) Resulting generated scheme.

Fig. A32. Setting up groups for the numeric interval and ratio attributes.